\documentclass[conference]{IEEEtran}
\IEEEoverridecommandlockouts
\usepackage{cite}
\usepackage{amsmath,amssymb,amsfonts}
\usepackage{algorithmic}
\usepackage{graphicx}
\usepackage{textcomp}
\usepackage{xcolor}
\def\BibTeX{{\rm B\kern-.05em{\sc i\kern-.025em b}\kern-.08em
    T\kern-.1667em\lower.7ex\hbox{E}\kern-.125emX}}
    
\usepackage{xcolor}
\usepackage{tikz}
\usepackage{subcaption}
\usepackage{soul}
\usepackage{booktabs}

\usepackage{mathtools}
\usepackage{amsmath, amsthm, amssymb, amsfonts, physics}

\DeclareMathSizes{10}{9}{6.5}{6.5}

\newtheorem{theorem}{Theorem}[section]

\newtheorem{lemma}{Lemma}[section]
\newtheorem{corollary}{Corollary}[section]
\newtheorem{definition}{Definition}[section]
\theoremstyle{definition}

\usepackage[square,numbers]{natbib}
\usepackage[hidelinks]{hyperref}

\begin{document}

\title{Resilience of Bayesian Layer-Wise Explanations \\ under Adversarial Attacks
}

\author{\IEEEauthorblockN{1\textsuperscript{st} Ginevra Carbone}
\IEEEauthorblockA{\textit{Dept. of Mathematics }\textit{and Geosciences}, \\
\textit{University of Trieste},\\
Trieste, Italy \\
ginevra.carbone@phd.units.it}
\and
\IEEEauthorblockN{2\textsuperscript{nd} Luca Bortolussi}
\IEEEauthorblockA{\textit{Dept. of Mathematics and Geosciences}, \\
\textit{University of Trieste},
Trieste, Italy \\
\textit{Modeling and Simulation Group}, \\
\textit{Saarland University},
Saarland, Germany \\
luca.bortolussi@gmail.com}
\and
\IEEEauthorblockN{3\textsuperscript{rd} Guido Sanguinetti}
\IEEEauthorblockA{\textit{School of Informatics},\\
\textit{University of Edinburgh},\\
Edinburgh, United Kingdom \\
\textit{Machine Learning and Systems}\\\textit{Biology group, SISSA},
Trieste, Italy \\
gsanguin@sissa.it }
}

\maketitle

\begin{abstract}
We consider the problem of the stability of saliency-based explanations of Neural Network predictions under adversarial attacks in a classification task. 
Saliency interpretations of deterministic Neural Networks are remarkably brittle even when the attacks fail, i.e. for attacks that do not change the classification label. We empirically show that interpretations provided by Bayesian Neural Networks are considerably more stable under adversarial perturbations of the inputs and even under direct attacks to the explanations. By leveraging recent results, we also provide a theoretical explanation of this result in terms of the geometry of the data manifold. Additionally, we discuss the stability of the interpretations of high level representations of the inputs in the internal layers of a Network. Our results demonstrate that Bayesian methods, in addition to being more robust to adversarial attacks, have the potential to provide more stable and interpretable assessments of Neural Network predictions.
\end{abstract}


\section{Introduction}

Deep Neural Networks (DNNs) are the core engine of the modern AI revolution. Their universal approximation capabilities, coupled with advances in hardware and training algorithms, have resulted in remarkably strong predictive performance on a variety of applications, from computer vision to natural language to bioinformatics. 

While the story of DNNs is indubitably one of success, it is tempered with a number of potentially very serious drawbacks which are somehow the natural flip side of  dealing with extremely flexible and complex models. The first such drawback is the black box nature of DNNs: their expressivity and training on large data sets empirically results in very strong predictive power, but in general it does not provide any intuition about the possible explanations underlying the decisions.
A second major drawback of DNN predictions is their vulnerability to adversarial attacks: empirically, it has been observed in many applications that well chosen infinitesimal changes in inputs can produce catastrophic changes in output \citep{goodfellow2015explaining}, leading to paradoxical classifications and a clear problem in any application to safety critical systems. Such brittleness appears to be intimately related to the geometry of the data itself \citep{carbone2020robustness}, and is therefore likely to be an intrinsic feature of standard DNN predictions.

In this paper, we argue theoretically and empirically \footnote{Code is available at \url{https://github.com/ginevracoal/BayesianRelevance}.}
that these two problems are interlinked, and that therefore solutions that ameliorate resilience against adversarial attacks will also lead to more stable and reliable interpretations.
We work within the framework of (pixel-wise) saliency explanations, which attempt to interpret post-hoc DNN decisions by apportioning a relevance score to each input feature for each data point.
Specifically, we use the popular Layer-wise Relevance Propagation (LRP) \cite{bach2015pixel}, a method to assess the contribution of each pixel to the final classification score which backpropagates the prediction in the neural network until it reaches the input, using a set of suitable propagation rules. LRP saliency interpretations are well known to be unstable under perturbations of the inputs \cite{ghorbani2019interpretation, kindermans2019reliability, alvarez2018towards, zhang2020interpretable}; recently, \cite{bykov2020much} suggested that a Bayesian treatment might ameliorate these stability problems. 

Here, we consider the stability of saliency interpretations under targeted adversarial attacks that aim to change the classification under perturbations of the input.  We introduce a novel notion of LRP robustness under adversarial attacks. As previously observed in \cite{ghorbani2019interpretation, heo2019fooling, dombrowski2019explanations}, our results confirm that the LRP robustness of deterministic DNN predictions is remarkably low even when the adversarial attack fail to change the overall classification of the data point, i.e. that LRP interpretations are {\it less robust} than actual classifications. Considerations on the geometry of LRP \citep{anders2020fairwashing} suggest that the observed lack of robustness might be imputable to large gradients of the prediction function in directions orthogonal to the data manifold. Here we expand on such a point of view, integrating it with a theoretical analysis in a suitably defined large-data limit \citep{carbone2020robustness, rotskoff2018neural, du2019gradient, mei2018mean}, and leveraging recent results from \cite{carbone2020robustness} about the robustness of BNNs to gradient based adversarial attacks. Specifically, we prove that Bayesian training of the DNNs in the large-data and overparameterized limit induces a regularizing effect which naturally builds robust explanations. We empirically validate this claim on the popular MNIST and Fashion MNIST benchmarks.

The main contributions of this paper are:
\begin{enumerate}
    \item The definition of a novel metric of robustness of interpretations of DNN results (Sec. \ref{sec:lrp_robustness});
    \item A theoretical analysis of the effects of adversarial attacks on the original inputs, the effects of saliency attacks on prediction interpretations, and  the improvements offered by a Bayesian treatment (Sec. \ref{sec:geometry});
    \item An empirical study showing that indeed Bayesian training and prediction leads to more robust interpretations of classifications (Sec. \ref{sec:experiments}).
\end{enumerate}
\section{Background}
\label{sec:background}

\subsection{Layer-Wise Relevance Propagation}\label{LRP_background}

Let $f:\mathbb{R}^d\rightarrow [0,1]$ be a single-class image classifier, where $f(x,w)$ is the probability that an image $x\in\mathbb{R}^d$  belongs to the class of interest and $w$ is the set of learnable weights. Without loss of generality, this concept can be extended to multi-class classifiers or to non probabilistic outputs. The idea of pixel-wise decomposition of a given image $x$ is to understand how its pixels contribute to the prediction $f(x,w)$. In particular, LRP associates with each pixel $p$ a relevance score $R(x_p,w)$, which is positive when the pixel  contributes positively to the classification, negative when it  contributes negatively to the classification and zero when it has no impact on the classification. 
All the relevance scores for a given image $x$ can be stored in a heatmap $R(x,w)=\{R(x_p,w)\}_p$, whose values quantitatively explain not only whether pixels contribute to the decision, but also by which extent. One can leverage suitable propagation rules to ensure that the network output is fully redistributed through the network, namely that the
relevance heatmap catches all the saliency features from the inputs \cite{samek2016evaluating}. For this purpose, the heatmap should be \emph{conservative}, i.e. the sum of the assigned relevances should correspond to the total relevance detected by the model:
$f(x,w)=\sum_p R(x_p,w).$
In the multi-label setting $R(x,w)$ is the heatmap associated to the classification label.
Although the conservative property is not required to define a relevance heatmap, it has been empirically observed that conservative rules better support classification \cite{samek2016evaluating, montavon2017explaining}. Several propagation rules satisfy the conservative property, each of them leading to different relevance measures. In Sec. \ref{sec:lrp_rules} of the supplementary material we report three practical propagation rules: the Epsilon rule, the Gamma rule and the Alpha-Beta rule.
\cite{bach2015pixel} also presented LRP using a functional approach, i.e. independently of the network’s topology. Then, \cite{montavon2017explaining} used \emph{deep Taylor decomposition} to express any rule-based approach under the functional setting. Their method builds on the standard first-order Taylor expansion of a non-linear classifier at a chosen \emph{root point} $x^*$, such that $f(x^*)=0$, 
\begin{align}
\begin{split}
    f(x)&=f(x^*)+\nabla_x f(x^*) \cdot (x-x^*)+\gamma\\
    &=\sum_p \frac{\partial f}{\partial x_p}\Big|_{x=x^*}\cdot(x_p-x^*_p)+\gamma,
\end{split}\label{LRP_gradients}
\end{align}
 where $\gamma$ denotes higher-order terms. The root point $x^*$ represents a neutral image which is similar to $x$, but does not influence classification, i.e. whose relevance is everywhere null. The nearest root point to the original image $x$ can be obtained by solving an iterative minimization problem \citep{montavon2017explaining}. The resulting LRP heatmap is $R(x)=\nabla_x f(x^*,w)\cdot(x-x^*)$. 
 
\subsection{Adversarial Attacks}

Adversarial attacks are small perturbations of input data that lead to large changes in output (in the classification case, changes in predicted label) \cite{goodfellow2015explaining}. Broadly speaking, most attack strategies utilise information on the DNN loss function to detect an optimal perturbation direction, either through explicit knowledge of the loss function ({\it white box attacks}) or via querying the DNN ({\it black box attacks}, generally weaker than their white box counterpart).

White box attacks generally utilize gradient information on the loss function to determine the attack direction. One of the best known gradient-based attacks is the Fast Gradient Sign Method (FGSM)~\citep{goodfellow2015explaining}, a one-step untargeted attack (i.e., an attack strategy which is independent of the class of the attacked point). An iterative, improved version of FGSM is Projected Gradient Descent (PGD) \cite{kurakin2017physical}. We define FGSM and PGD in Section \ref{sec:gradient_attacks} of the Appendix.

The evaluation of adversarial attacks can be either qualitative or quantitative. In the qualitative case, one simply observes whether an attack strategy with a certain strength is successful in switching the classification label of the given data point; in general, the evaluation then reports the fraction of successful attacks on the whole data set. A quantitative metric to evaluate network performances against adversarial attacks
is provided by the notion of \emph{softmax robustness} \citep{carbone2020robustness}, which computes the softmax difference between original and adversarial predictions as $1-||f(x)-f(\tilde{x})||_\infty$
and is a number between zero (maximal fragility) and one (complete robustness) for every data point.

\subsection{Bayesian Neural Networks}\label{BNN_background}
Bayesian models capture the inherent uncertainty intrinsic in any model by replacing individual models with ensembles weighted by probability laws. In the DNN setup, a
Bayesian Neural Network (BNN) consists of an ensemble of DNNs whose weights are drawn from the posterior measure $p(w\vert D)$, where $D$ denotes the available training data. To compute the posterior measure, one needs to first define a prior distribution $p(w)$ on the network's weights $w$; by Bayes' theorem, the posterior is then obtained by combining the prior and the likelihood $p(w\vert D)\propto p(D \vert w)p(w)$, where the likelihood term $p(D\vert w)$ quantifies the fit of the network with weights $w$ to the available training data. Exact computation of the posterior distribution $p(w\vert D)$ is in general infeasible, thus one needs to resort to approximate inference methods for training Bayesian NNs, namely Hamiltonian Monte Carlo (HMC)~\citep{neal2011mcmc} and Variational Inference (VI)~\citep{wainwright2008graphical}. We describe both techniques in Section \ref{sec:approximate_inference} of the appendix.

Once the posterior distribution has been computed, BNNs produce predictions through the {\it posterior predictive distribution} 
\begin{align*}
    p(f(x)\vert D)
    &=\int_w p(f(x)\vert w)p(w\vert D)dw\\
    &\simeq \sum_{w_i\sim p(w\vert D)}p(f(x)\vert w_i),
\end{align*}
where $f(x)$ is the output value at a new point $x$ and the equality represents the celebrated {\it Bayesian model averaging} procedure.
In the BNN setting, adversarial attacks are crafted against the posterior predictive distribution $p(f(x)\vert D)$. For example, FGSM attack on an ensemble of $N$ networks $f(\cdot,w_i)$ with weights drawn from $p(w\vert D)$ becomes
\begin{align*}
    \tilde{x}
    &= x+\delta \; \text{sgn} \Big(\mathbb{E}_{p(w\vert D)}\big[\nabla_x L(x,w)\big]\Big)\\
    &\simeq x + \delta \; \text{sgn}  \sum_{i=1}^N\nabla_x L(x,w_i) \qquad w_i\sim p(w\vert D).
\end{align*}

In a similar way, it is possible to introduce a concept of {\it Bayesian explanations} for BNN predictions. As is clear from the definition(s) of LRP in Section \ref{LRP_background}, the relevance score assigned to an input feature depends on the neural network weights. Therefore, in the Bayesian setting, the relevance becomes a random variable; we define as a Bayesian explanation the expected relevance of a feature under the posterior distribution of the weights.

\begin{figure}[hb]
	\centering
	\includegraphics[width=0.7\linewidth]{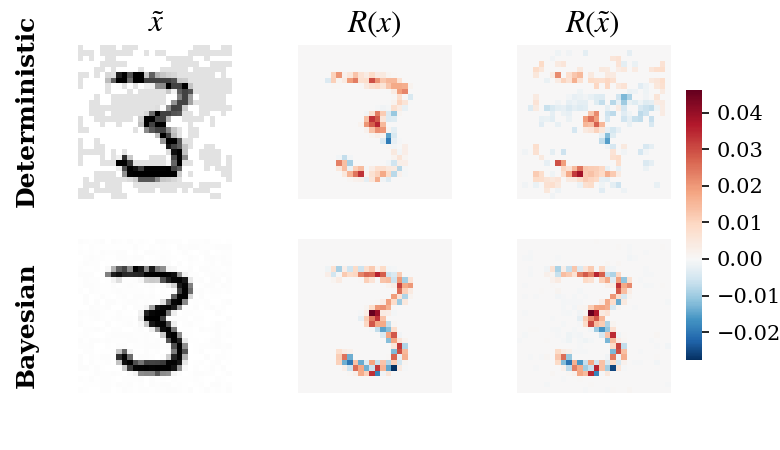}
	\caption{LRP heatmaps of an image $x$ (second column) and an FGSM adversarial perturbation $\tilde{x}$ (third column) which fails on a deterministic network and a Bayesian network. The two models have the same fully connected architecture and are both trained on the MNIST dataset. Explanations are computed using the Epsilon rule on the pre-softmax layer, i.e. layer idx $=5$. Bayesian LRP is computed using $100$ posterior samples. For each heatmap we only show the $30\%$ most relevant pixels. The LRP robustness amounts to $0.46$ in the deterministic case and to $0.68$ in the Bayesian case.
	}
	\label{fig:heatmaps_det_vs_bay}
\end{figure}

Recent research shows that Bayesian neural networks are adversarially robust to gradient-based attacks in the overparameterized and infinite data limit ~\citep{carbone2020robustness, rotskoff2018neural}. It is therefore of interest to  investigate whether such robustness also extends to the learned explanations.
To do so, we compare the explanations of deterministic NNs to that of Bayesian NNs against such attacks. Fig. \ref{fig:heatmaps_det_vs_bay} shows an example of failed FGSM attack for a deterministic network and a Bayesian network with the same architecture. Despite the fact that the attack did not manage to change the overall classification, we can see immediately a large difference between the deterministic LRP explanation of the original image, $R(x)$, and of the adversarial image, $R(\tilde{x})$ (top row). On the other hand, in the Bayesian case (bottom row), the saliency maps before and after the attack are essentially identical.  In the next sections, we will provide both a theoretical explanation of this phenomenon, and systematically substantiate empirically the robustness of Bayesian explanations to adversarial attacks.

\section{Related Work}

Most of the recent explanation methods provide post hoc interpretations of black box classifier, which generate visual explanations of the decisions performed on single input samples \cite{adadi2018peeking, guidotti2018survey, marcinkevivcs2020interpretability}. 
Among the variety of available techniques, gradient-based attribution methods rely on gradient information provided by the models to produce the explanations. We briefly mention a few of them in what follows. \cite{simonyan2013deep} compute image-specific saliency maps using a single back-propagation step across the network for a chosen class.
\emph{Local Interpretable Model-agnostic Explanations} (LIME) \cite{ribeiro2016should} searches for the optimal explanation of a sample from a specified class of explanation models which are intrinsically interpretable \cite{marcinkevivcs2020interpretability}. The attribution method of
\emph{Integrated Gradients} \cite{sundararajan2017axiomatic} satisfies two specific axioms: \emph{sensitivity} and \emph{implementation invariance}. Sensitivity indicates that whenever an input and a baseline differ by a single feature and their predictions on that input are distinct, the attribution associated to the differing feature is non-zero. Two functionally equivalent networks satisfy implementation invariance if they associate identical attributions to the same input. Integrated gradients have been extended in several works \cite{dhamdhere2018important, jha2020enhanced, merrill2019generalized}.
\emph{Deep learning important features} (DeepLIFT) \cite{shrikumar2017learning}
assigns attributions by comparing the activation of each neuron to a reference activation, i.e. the activation of a baseline input, which is task-dependent. It satisfies the sensivity axiom but breakes implementation invariance \cite{sundararajan2017axiomatic}. \emph{Shapley additive explanations} (SHAP) \cite{lundberg2017unified} generalizes all the explanation functions that can be expressed as a linear combination of binary variables, including LIME and DeepLIFT.

Adversarial machine learning has been extensively studied from the perspective of interpretability. Many works focus on adversarial manipulations of the explanations \cite{heo2019fooling, ghorbani2019interpretation, dombrowski2019explanations}, which consists in altering the explanations without changing model predictions. Interestingly, \cite{rieger2020simple} notice that aggregations of multiple explanation methods are less vulnerable to adversarial attacks. It has been observed that there is a strong connection between adversarial robustness and explainability of Neural Networks \cite{tomsett2018failure}. For instance, \cite{dong2017towards} use an adversarial training \cite{goodfellow2015explaining, madry2017towards} to improve the interpretability of the representations, while \cite{ross2018improving} present a regularization technique based on gradient smoothing, which favours adversarial robustness and interpretability simultaneously. Adversarial training is one of the most effective defense methods against the attacks, but it has also been used to improve sensitivity of the explanations to input perturbations \cite{yeh2019fidelity}. In contrast to the previous studies, we provide a new concept of robustness of the explanations, where we compare the interpretation of an input to that of an adversarial attack in terms of common ``relevant'' pixels (Sec. \ref{sec:methodology}). 

A recent work \cite{bykov2021noisegrad} investigates how the stability of the interpretations could be improved by adding stochasticity to the model weights. Their \emph{NoiseGrad} method relies on a \emph{tempered Bayes posterior} \cite{wenzel2020good} and aggregates the explanations provided by independent samples to produce a Bayesian explanation. Our idea that Bayesian Neural Networks could provide more stable explanations, instead, is motivated by the adversarial robustness of BNNs to gradient-based attacks \cite{carbone2020robustness}. Hence, we focus on gradient-based attacks and also provide a formal proof for the stability of LRP explanations (Thm. \ref{th:main_thm}), which holds w.l.o.g. for any other gradient-based attribution method. Furthermore, we empirically evaluate our findings using VI and HMC approximate inference methods (Sec. \ref{sec:experiments}).

\section{Methodology}
\label{sec:methodology}

\subsection{LRP Robustness}
\label{sec:lrp_robustness}

We define the \emph{$k$-LRP robustness} of relevance heatmaps to adversarial attacks and use this measure to assess how adversarial perturbations of the inputs affect the explanations.

\begin{definition}
Let $x$ be an image with relevance heatmap $R(x, w)$ and $\tilde{x}$ an adversarial perturbation with relevance heatmap $R(\tilde{x}, w)$. Let $\text{Top}_k(R)$ denote the pixel indexes corresponding to the top $k$ percent most relevant pixels in the absolute value of a heatmap $R$. 
The \emph{$k$-LRP robustness} of $x$ w.r.t. the attack $\tilde{x}$ is
\begin{align}
k\text{-LRP}(x,\tilde{x},w):=\frac{|\text{Top}_k(R(x, w))\cap\text{Top}_k(R(\tilde{x}, w))|}{k}.
\label{eq:lrp_robustness}
\end{align}
\end{definition}

In other words, the $\text{Top}_k(R)$ pixels have a strong positive or negative impact on classification and $k\text{-LRP}(x,\tilde{x},w)$ is the fraction of common  most relevant pixels for $x$ and $\tilde{x}$ in the top $k\%$. Fig. \ref{fig:topk_rob} in the Appendix gives an intuition of this computation.
Notice that the LRP robustness of a point depends only implicitly on the strength of the attack through the attacked point $\tilde{x}$.

\paragraph{Inner layers explanations}

\begin{figure}[bt]
	\centering
	\includegraphics[width=0.95\linewidth]{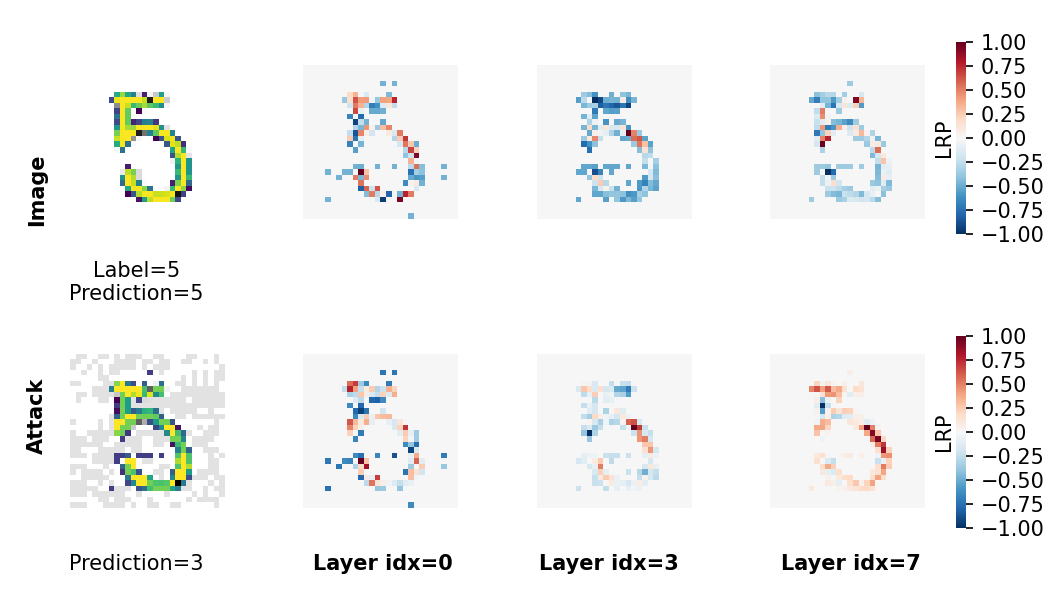}
	\caption{
	LRP heatmaps of an image $x$ and an FGSM adversarial perturbation $\tilde{x}$. Explanations are computed using the Epsilon rule w.r.t. the learnable layers of a deterministic network trained on the MNIST dataset. For each layer, we show the $30\%$ most relevant pixels for both the original image and its adversarial counterpart, i.e. the ones selected for the computation of $k\text{-LRP}(x, \tilde{x},w,l)$ from Eq. \eqref{eq:lrp_robustness}. 
	}
	\label{fig:heatmaps}
\end{figure}

We analyse the behaviour of LRP representations in the internal layers of the network, thus we also extend the computation of LRP heatmaps to any feature representation of the input $x$ at a learnable layer $l\in\mathbb{N}$. We denote it by $R(x,w,l)$, where $l\leq L$ and $L$ is the maximum number of layers available in the architecture. The corresponding LRP robustness will be denoted by $k\text{-LRP}(x,\tilde{x},w,l)$.
In such case, the robustness does not refer anymore to explanations in the classification phase (pre-softmax layer), but rather to the explanations in the learning phases, hence it gives an idea of the most relevant pixels determining an internal representation. 

Fig. \ref{fig:heatmaps} shows an example of internal LRP heatmaps on a deterministic NN with learnable layers indexed by $l\in[0,3,7]$. For illustrative purposes, heatmaps appearing on the same row are normalized in $[-1,1]$ before selecting the $\text{Top}_k$ pixels, since numeric scales are significantly different across the different internal representations.

\paragraph{Bayesian LRP robustness}
The notion of LRP robustness can be naturally generalised to the Bayesian setting using the concept of Bayesian model averaging introduced in Section \ref{BNN_background}.
Hence, the LRP heatmap of a BNN is computed as the average of all the deterministic heatmaps from the ensemble:
$\mathbb{E}_{p(w|D)}\big[k\text{-LRP}(x,\tilde{x},w,l)\big]$.
In this regard, we emphasise that Bayesian interpretations are affected by the chosen number of posterior samples drawn from the learned distribution.

\subsection{Geometric meaning of adversarial interpretations}
\label{sec:geometry}
To better conceptualise the impact of a Bayesian treatment on LRP robustness, it is convenient to consider the thermodynamic limit of infinite data and infinite expressivity of the network, as formalised in \cite{du2019gradient,mei2018mean,rotskoff2018neural}. 
For the purposes of our discussion, the main ingredients are the data manifold $\mathcal{M}_D$, a piecewise smooth submanifold of the input space where the data lie, and the true input/output function, which is assumed to be smooth and hence representable through an infinitely wide DNN. Practically, this limit might be well approximated on large data sets where the networks achieve high accuracy.

In this limit, it is proved \cite{du2019gradient,mei2018mean,rotskoff2018neural} that the DNN $f(x,w)$ trained via SGD will converge to the true underlying function $g(x)$ over the whole data manifold $\mathcal{M}_D$. Because the data manifold is assumed to be piecewise smooth, it is possible to define a tangent space to the data manifold almost everywhere, and therefore to define two operators $\nabla_x^{\perp}$ and $\nabla_x^{\parallel}$ which define the gradient along the normal and tangent directions to the data manifold $\mathcal{M}_D$ at a point $x$ of a function defined over the whole input space. 

LRP and gradient-based adversarial strategies both share a reliance on gradient information. In the adversarial attacks cases, one evaluates the gradient of the loss function which, by the chain rule is given by
\begin{equation}
    \nabla_x L(f,g)=\frac{\delta L(f,g)}{\delta f}\frac{\partial f}{\partial x} \label{loss_gradient}
\end{equation}
In the thermodynamic limit, the DNN function $f(x,w)$ coincides with the true function everywhere on the data manifold, and therefore the tangent gradient of the loss function is identically zero. The normal gradient of the loss, however, is unconstrained by the data, and, particularly in a high dimensional setting, might achieve very high values along certain directions, creating therefore weaknesses that may be exploited by an adversarial attacker. The main result in \cite{carbone2020robustness} was to show that the orthogonal component of the loss gradient has expectation zero under the posterior weight distribution, therefore showing that BNNs are robust against adversarial attacks.

In the LRP setup, we instead consider gradients of the prediction function, as opposed to the loss, nevertheless the insight remains valid. The tangent components of the gradient of the prediction function $f(x,w)$ will coincide with the gradients of the true function $g(x)$, and therefore represent directions of true sensitivity of the decision function which are correctly recognised as relevant. However, such directions might be confounded or dwarfed by normal gradient components, which create directions of apparent relevance which, by construction, are targeted by gradient-based adversarial attacks. In Theorem \ref{th:main_thm} we prove that BNNs in the thermodynamic limit will only retain relevant directions along the data manifold, which correspond to genuine directions of high relevance. 

\begin{theorem}\label{th:main_thm}
Let $\mathcal{M}_D\subset \mathbb{R}^d$ be an a.e. smooth data manifold and let $f(x,w)$ be an infinitely wide Bayesian neural network, trained on $\mathcal{M}_D$ and at full convergence of the training algorithm. Let $p(w|D)$ be the posterior weight distribution and suppose that the prior distribution $p(w)$ is uninformative. In the limit of infinite training data, for any $x\in\mathcal{M}_D$, 
$$
\mathbb{E}_{p(w|D)}[\nabla_x^\perp f(x,w)]=0.
$$
\end{theorem}

Therefore, the orthogonal component of the gradient of the prediction function vanishes in expectation under the posterior weight distribution and Bayesian averaging of the relevance heatmaps naturally builds explanations in the tangent space $T_x\mathcal{M}_D$. Specifically, from the deep Taylor decomposition on a root point $x^*\in\mathcal{M}_D$ we obtain the expected LRP heatmap 
\begin{align*}
\mathbb{E}_{p(w|D)}[R(x)] 
&= \mathbb{E}_{p(w|D)}\big[\nabla^{\parallel}_x f(x^*,w)\big]\cdot(x-x^*).
\end{align*}

We refer the reader to Sec. \ref{sec:theoretical_results} of the Appendix for the theoretical background material and for a formal proof of Theorem \ref{th:main_thm}.
It should be noticed that LRP heatmaps at layer $l$ involve partial derivatives w.r.t. $x$ of the subnetwork $f^l(\cdot,w)$ of $f$, which associates to an input $x$ the $l$-th activation from $f(\cdot,w)$. Consequently, the same vanishing property of the gradients holds for explanations in the internal layers - which are therefore more robust in the Bayesian case.
\section{Experimental Results}
\label{sec:experiments}
In this section we corroborate the insights described in Section \ref{sec:methodology} with an experimental evaluation, comparing empirically the
LRP robustness using the popular MNIST~\citep{mnist} and Fashion MNIST~\citep{xiao2017fashion} benchmark data sets. Both data sets are composed of $60.000$ training images belonging to ten classes: in the MNIST case, these are hand-written digits, while the Fashion MNIST data set consists of stylized Zalando images of clothing items. While MNIST is considered a relatively trivial data set, with accuracies over 99\% being regularly reported, Fashion MNIST is considerably more complex, and the best architectures report accuracies around 95\%. We also extend the experiments to a ResNet architecture trained on 3-channels images from CIFAR-10 \cite{cifar10}, which consists of $50.000$ 3-channels training images from ten classes. We do not examine more complex data sets, such as ImageNet \citep{imagenet_cvpr09}, because of the very high computational costs of running Bayesian inference on very deep networks trained on very large data sets \footnote{We do not experiment with scalable Monte Carlo dropout methods \cite{gal2016dropout} here since there is no guarantee that their uncertainty estimates are able to capture the full posterior \cite{smith2018understanding}.}. 
We train multiple DNNs and BNNs using both HMC and VI, which allows us to contrast the effect of a locally Gaussian approximation to the posterior against the asymptotically exact (but computationally more expensive) approximation provided by HMC. Because we require high accuracy in order to approximate the asymptotic conditions described in Section \ref{sec:geometry}, different architectures were used on the three data sets and between VI and HMC. In all cases, however, the BNN is compared with a DNN with the same architecture, to ensure fairness of the comparisons.
Full details of the architectures used are reported in Sec. \ref{sec:architectures} of the Appendix. 
Adversarial attacks in our tests are
Fast Gradient Sign Method
and Projected Gradient Descent, 
with a maximum perturbation size of $0.2$. Saliency attacks are \emph{target region} and \emph{top-k} attacks \cite{ghorbani2019interpretation} and \emph{beta} attacks \cite{dombrowski2019explanations}, which we briefly describe in Sec. \ref{sec:saliency_attacks} of the Appendix.
We rely on \emph{TorchLRP} library \footnote{\url{https://github.com/fhvilshoj/TorchLRP}} for the computation of LRP explanations and set $\epsilon=0.1$ in the Epsilon rule, $\gamma=0.1$ in the Gamma rule, $\alpha=1$ and $\beta=0$ in the Alpha-Beta rule. We report the computational resources in Sec. \ref{sec:comput_resources} of the Appendix.

\subsection{Bayesian interpretations are robust against the attacks}

\begin{figure}[bt]
	\centering
	\begin{subfigure}{0.48\linewidth}
        \centering	\includegraphics[width=\linewidth]{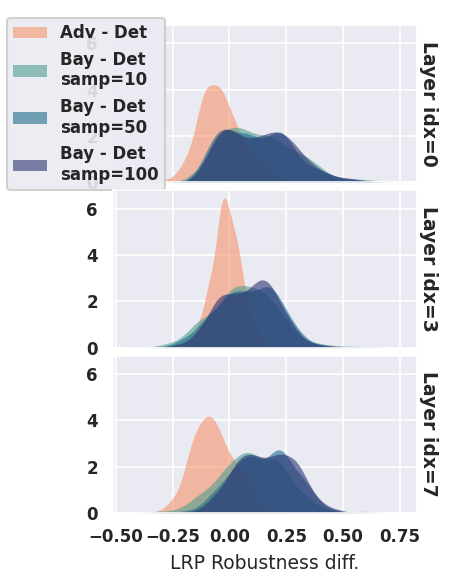}
        \caption{}
    \end{subfigure}%
    \hfill
	\begin{subfigure}{0.5\linewidth}
    	\centering	\includegraphics[width=\linewidth]{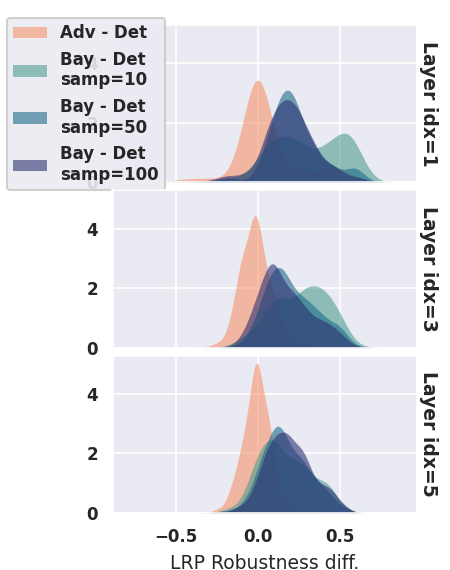}
    	\caption{}
    \end{subfigure}
    \caption{LRP robustness differences for FGSM (a) and PGD (b) attacks computed on $500$ test points from Fashion MNIST dataset using the Epsilon rule. NNs in (a) have a convolutional architecture (Tab. \ref{tab:conv_architecture} in the Appendix)  and the BNN is trained with VI. NNs in (b) have a fully connected architecture (Tab. \ref{tab:fc_architecture} in the Appendix)
    and the BNN is trained with HMC.  BNNs are tested using an increasing number of samples ($10,50,100$). Layer indexes refer to the learnable layers in the architectures.}
    \label{fig:robusteness_distr_main}
\end{figure}

Our first significant result is that Bayesian explanations are more robust under attacks than deterministic architectures. For multiple data sets, attacks, training techniques (deterministic training, adversarial training, Bayesian inference) and approximate inference methods, LRP robustness scores are considerably higher than their deterministic counterparts. In Fig. \ref{fig:robusteness_distr_main} and Fig. \ref{fig:rob_diff_appendix_1},\ref{fig:rob_diff_appendix_2},\ref{fig:rob_diff_appendix_3} in the Appendix we show the distribution of point-wise differences w.r.t. the deterministic baselines between LRP robustness scores for MNIST and Fashion MNIST data sets using the Epsilon rule.
The bottom row of the figures is the standard LRP (computed from the pre-softmax layer), while the top row is the initial feature representation (after the first non-linear layer), and the middle row is the LRP of an internal layer.
We tested Bayesian representations using an increasing number of posterior samples, i.e. $10, 50, 100$. We attacked $500$ randomly selected test images, whose choice is balanced w.r.t. the available classes.

The first notable observation is that adversarially trained networks have low LRP robustness compared to BNNs: this confirms empirically the conjecture of Section \ref{sec:geometry} that the components of the gradient that are normal to the data manifold (and are therefore the ones likely to be changed in an attack) are often major contributors to the relevance in DNN. On the contrary, the Bayesian averaging process greatly reduces the expected relevance of such direction. 


\begin{figure}[bt]
	\centering	\includegraphics[width=0.8\linewidth]{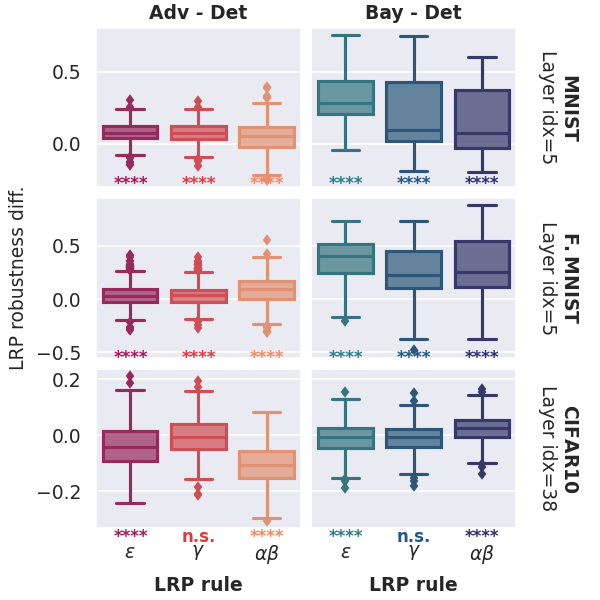}
	\caption{LRP robustness differences for FGSM attack attacks computed on $500$ test points from MNIST  (top row), Fashion MNIST (middle) and CIFAR-10 datasets (bottom) using Epsilon, Gamma and Alpha-Beta rules (columns) on the $\text{Top}_{20}$ pixels, for adversarially trained networks (left block) and BNNs (right block). NNs for MNIST and Fashion MNIST have a fully connected architecture (Tab. \ref{tab:fc_architecture}) and the BNNs are trained with HMC. NNs for CIFAR-10 have a ResNet20 architecture from \texttt{bayesian\_torch} library \cite{krishnan2020bayesiantorch} and the BNN is trained with VI. BNNs are tested using $100$ posterior samples. Layer indexes refer to the last learnable layer in each architecture. Parameters are described in the main text.}
	\label{fig:robustness_rules}
\end{figure}

In Fig. \ref{fig:robustness_rules} and \ref{fig:rules_rob_appendix_1}-\ref{fig:rules_rob_appendix_cifar} in the supplementary material we test the stability of deterministic and Bayesian interpretations w.r.t. FGSM and PGD attacks on $500$ test inputs from MNIST, Fashion MNIST and CIFAR-10 datasets. In Fig. \ref{fig:rules_rob_appendix_ghorbani} and Fig. \ref{fig:rules_rob_appendix_dombrowski} in the Appendix, we also test LRP stability against top-k, target region and beta saliency attacks (Sec. \ref{sec:saliency_attacks}).  As the saliency attacks are specifically designed to harm the interpretations, we consider them as proxy for the worst case robustness of the interpretations in a neighbor of the attacked point. Additionally, in Sec. \ref{sec:additional_figures} of the Appendix we compare the robustness distributions for adversarially trained and Bayesian NNs using Mann-Whitney U rank test.
We compute the relevance heatmaps using Gamma, Epsilon and Alpha-Beta rules. BNNs are trained with HMC and VI and evaluated using $100$ samples from the posterior. The experiments confirm that Bayesian explanations are more stable across multiple LRP rules, gradient-based adversarial attacks and saliency attacks, also in the internal layers.
Experiments on CIFAR-10 images in Fig. \ref{fig:robustness_rules} (bottom row) with a ResNet architecture from \texttt{bayesian\_torch} library \cite{krishnan2020bayesiantorch} show only a modest improvement in adversarial and LRP robustness, probably due to the conditions of Thm \ref{th:main_thm} not being satisfied in this data set with higher complexity but relatively small size.


\begin{figure}[bt]
    \centering
    	\begin{subfigure}{0.5\linewidth}
            \centering	\includegraphics[width=\linewidth]{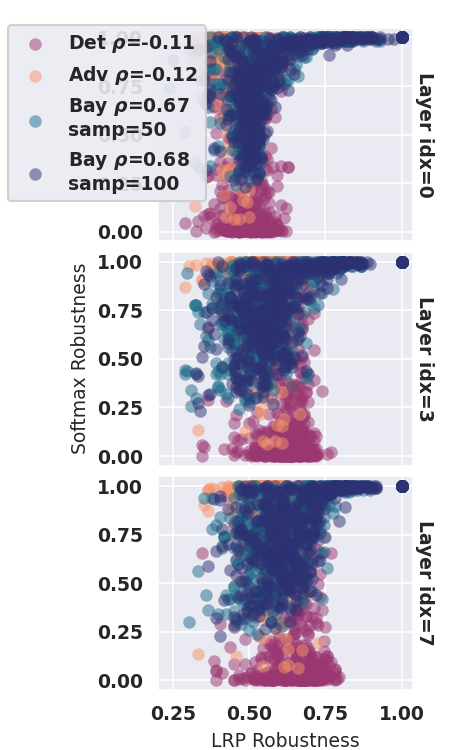}
            \caption{}
        \end{subfigure}%
        \hfill
    	\begin{subfigure}{0.48\linewidth}
        	\centering	\includegraphics[width=\linewidth]{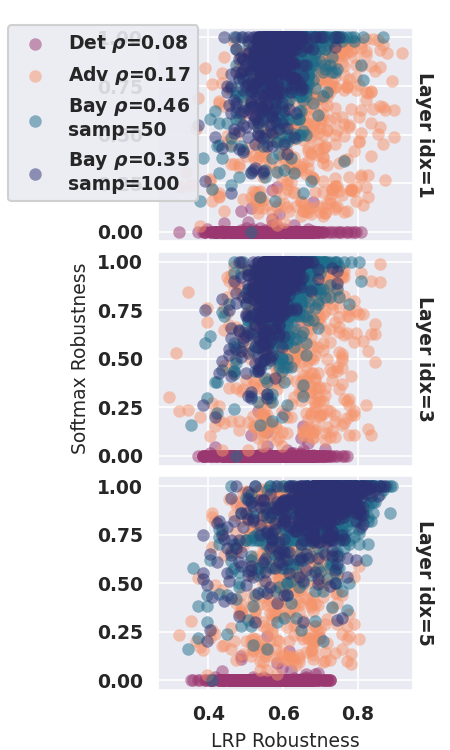}
        	\caption{}
        \end{subfigure}
	\caption{LRP vs softmax robustness of deterministic, adversarially trained and Bayesian NNs trained on MNIST dataset and tested against FGSM (a) and PGD (b) attacks. $\rho$ denotes Spearman correlation coefficient. LRP Robustness is computed with the Epsilon rule on the $20\%$ most relevant pixels. BNNs are trained with VI (a) and HMC (b) and are tested using an increasing number of samples ($50,100$). Layer indexes refer to the learnable layers in the convolutional (a) and fully connected (b) architectures (Tab. \ref{tab:conv_architecture} and \ref{tab:fc_architecture} in the Appendix).
	}
	\label{fig:scatterplot_main}
\end{figure}

\subsection{Bayesian LRP robustness increases with softmax robustness}
A simple explanation for the improved LRP robustness of BNNs lies in the fact that BNNs are provably immune to gradient-based attacks \citep{carbone2020robustness}. Therefore, one might argue that the stability of the LRP is a trivial byproduct of the stability of the classifications.

To explore this question more in depth, we consider the relationship between the LRP robustness of a test point (stability of the explanation) and its softmax robustness (resilience of the classification against an attack).
Fig. \ref{fig:scatterplot_main} and Fig. \ref{fig:scatterplot_appendix} in the supplementary material show scatterplots of these two quantities. An immediate observation is that deterministic explanations are weak against adversarial perturbations even when their softmax robustness 
is close to $1$.
Therefore, even in the cases where the classification is unchanged, deterministic saliency heatmaps are fragile. In fact, there are no significant changes in LRP robustness between data points that are vulnerable to attacks and data points that are robust to attacks. Bayesian models, instead, show a strong positive correlation between LRP and softmax robustness, especially as the number of posterior samples increases. While it is immediately evident that Bayesian predictions are robust to adversarial attacks (since most data points have softmax robustness greater than 0.5), it is also clear from this correlation that attacks which are more successful (i.e. lower softmax robustness) also alter more substantially the interpretation, and are likely to represent genuine directions of change of the true underlying decision function along the data manifold. 

\section{Conclusions}
\label{sec:conclusions}

Deep neural networks are fundamental to modern AI, yet many of the mathematical structures underlying their success are still poorly understood. Unfortunately, an unavoidable consequence of this situation is that we also lack principled tools to address the weaknesses of deep learning systems.
In this paper, we harness the geometric perspective to adversarial attacks introduced in \cite{carbone2020robustness} to study the resilience of Layer-wise Relevance Propagation heatmaps to adversarial attacks. The geometric analysis suggests a fundamental link between the fragility of DNNs against adversarial attacks and the difficulties in understanding their predictions: because of the unconstrained nature of classifiers defined on high dimensional input spaces but trained on low dimensional data, gradients of both the loss function and the prediction function tend to be dominated by directions which are orthogonal to the data manifold. These directions both give rise to adversarial attacks, and provide spurious explanations which are orthogonal to the natural parametrization of the data manifold. 
In the limit of infinite data, a Bayesian treatment remedies the situation by averaging out irrelevant gradient directions in expectation. 
Not only BNN interpretations are considerably more robust than deterministic DNN, but we also observe a correlation between softmax (adversarial) robustness and LRP robustness which suggests that indeed Bayesian interpretations  capture the relevant parametrization of the data manifold. 

We point out the presence of theoretical and practical limitations. The strong assumptions in Theorem \ref{th:main_thm}, which restrict the geometrical considerations to fully trained BNNs in the limit of an infinite amount of weights and training data, do not prevent us from observing the desired behavior in practice, even when using cheap approximate inference techniques (VI). Indeed, Bayesian interpretations are considerably more robust than deterministic and adversarially trained networks.
However, learning accurate BNNs on more complex datasets is extremely challenging, which makes the Bayesian scheme currently not suitable for large-scale applications. 
This suggests the need for further investigations on such matters,  especially on sufficiently accurate and scalable approximate inference methods for BNNs
\cite{wenzel2020good}.
Nevertheless, we believe that the insights provided by a geometric interpretation will be helpful towards a better understanding of both the strengths and the weaknesses of deep learning.

\bibliographystyle{IEEEtran}
\bibliography{bibliography}

\begin{thebibliography}{10}
\providecommand{\url}[1]{#1}
\csname url@samestyle\endcsname
\providecommand{\newblock}{\relax}
\providecommand{\bibinfo}[2]{#2}
\providecommand{\BIBentrySTDinterwordspacing}{\spaceskip=0pt\relax}
\providecommand{\BIBentryALTinterwordstretchfactor}{4}
\providecommand{\BIBentryALTinterwordspacing}{\spaceskip=\fontdimen2\font plus
\BIBentryALTinterwordstretchfactor\fontdimen3\font minus
  \fontdimen4\font\relax}
\providecommand{\BIBforeignlanguage}[2]{{%
\expandafter\ifx\csname l@#1\endcsname\relax
\typeout{** WARNING: IEEEtran.bst: No hyphenation pattern has been}%
\typeout{** loaded for the language `#1'. Using the pattern for}%
\typeout{** the default language instead.}%
\else
\language=\csname l@#1\endcsname
\fi
#2}}
\providecommand{\BIBdecl}{\relax}
\BIBdecl

\bibitem{goodfellow2015explaining}
\BIBentryALTinterwordspacing
I.~Goodfellow, J.~Shlens, and C.~Szegedy, ``Explaining and harnessing
  adversarial examples,'' \emph{ICLR}, 2015. [Online]. Available:
  \url{http://arxiv.org/abs/1412.6572}
\BIBentrySTDinterwordspacing

\bibitem{carbone2020robustness}
G.~Carbone, M.~Wicker, L.~Laurenti, A.~Patane, L.~Bortolussi, and
  G.~Sanguinetti, ``Robustness of bayesian neural networks to gradient-based
  attacks,'' in \emph{NeurIPS}, vol.~33.\hskip 1em plus 0.5em minus 0.4em\relax
  Curran Associates, Inc., 2020, pp. 15\,602--15\,613.

\bibitem{bach2015pixel}
S.~Bach, A.~Binder, G.~Montavon, F.~Klauschen, K.-R. M{\"u}ller, and W.~Samek,
  ``On pixel-wise explanations for non-linear classifier decisions by
  layer-wise relevance propagation,'' \emph{PloS one}, vol.~10, no.~7, p.
  e0130140, 2015.

\bibitem{ghorbani2019interpretation}
A.~Ghorbani, A.~Abid, and J.~Zou, ``Interpretation of neural networks is
  fragile,'' in \emph{Proceedings of AAAI}, vol.~33, no.~01, 2019, pp.
  3681--3688.

\bibitem{kindermans2019reliability}
P.-J. Kindermans, S.~Hooker, J.~Adebayo, M.~Alber, K.~T. Sch{\"u}tt,
  S.~D{\"a}hne, D.~Erhan, and B.~Kim, ``The (un) reliability of saliency
  methods,'' in \emph{Explainable AI: Interpreting, Explaining and Visualizing
  Deep Learning}.\hskip 1em plus 0.5em minus 0.4em\relax Springer, 2019, pp.
  267--280.

\bibitem{alvarez2018towards}
D.~Alvarez-Melis and T.~S. Jaakkola, ``Towards robust interpretability with
  self-explaining neural networks,'' \emph{arXiv preprint arXiv:1806.07538},
  2018.

\bibitem{zhang2020interpretable}
X.~Zhang, N.~Wang, H.~Shen, S.~Ji, X.~Luo, and T.~Wang, ``Interpretable deep
  learning under fire,'' in \emph{29th USENIX}, 2020.

\bibitem{bykov2020much}
K.~Bykov, M.~M.-C. H{\"o}hne, K.-R. M{\"u}ller, S.~Nakajima, and M.~Kloft,
  ``How much can i trust you?--quantifying uncertainties in explaining neural
  networks,'' \emph{arXiv preprint arXiv:2006.09000}, 2020.

\bibitem{heo2019fooling}
J.~Heo, S.~Joo, and T.~Moon, ``Fooling neural network interpretations via
  adversarial model manipulation,'' \emph{arXiv preprint arXiv:1902.02041},
  2019.

\bibitem{dombrowski2019explanations}
A.-K. Dombrowski, M.~Alber, C.~J. Anders, M.~Ackermann, K.-R. M{\"u}ller, and
  P.~Kessel, ``Explanations can be manipulated and geometry is to blame,''
  \emph{arXiv preprint arXiv:1906.07983}, 2019.

\bibitem{anders2020fairwashing}
C.~Anders, P.~Pasliev, A.-K. Dombrowski, K.-R. M{\"u}ller, and P.~Kessel,
  ``Fairwashing explanations with off-manifold detergent,'' in
  \emph{ICML}.\hskip 1em plus 0.5em minus 0.4em\relax PMLR, 2020, pp. 314--323.

\bibitem{rotskoff2018neural}
G.~M. Rotskoff and E.~Vanden-Eijnden, ``Neural networks as interacting particle
  systems: Asymptotic convexity of the loss landscape and universal scaling of
  the approximation error,'' \emph{stat}, vol. 1050, p.~22, 2018.

\bibitem{du2019gradient}
S.~Du, J.~Lee, H.~Li, L.~Wang, and X.~Zhai, ``Gradient descent finds global
  minima of deep neural networks,'' in \emph{ICML}.\hskip 1em plus 0.5em minus
  0.4em\relax PMLR, 2019, pp. 1675--1685.

\bibitem{mei2018mean}
S.~Mei, A.~Montanari, and P.-M. Nguyen, ``A mean field view of the landscape of
  two-layer neural networks,'' \emph{PNAS}, vol. 115, no.~33, pp. E7665--E7671,
  2018.

\bibitem{samek2016evaluating}
W.~Samek, A.~Binder, G.~Montavon, S.~Lapuschkin, and K.-R. M{\"u}ller,
  ``Evaluating the visualization of what a deep neural network has learned,''
  \emph{IEEE transactions on neural networks and learning systems}, vol.~28,
  no.~11, pp. 2660--2673, 2016.

\bibitem{montavon2017explaining}
G.~Montavon, S.~Lapuschkin, A.~Binder, W.~Samek, and K.-R. M{\"u}ller,
  ``Explaining nonlinear classification decisions with deep taylor
  decomposition,'' \emph{Pattern Recognition}, vol.~65, pp. 211--222, 2017.

\bibitem{kurakin2017physical}
A.~Kurakin, I.~J. Goodfellow, and S.~Bengio, ``Adversarial examples in the
  physical world,'' \emph{CoRR}, vol. abs/1607.02533, 2016.

\bibitem{neal2011mcmc}
R.~M. Neal \emph{et~al.}, ``Mcmc using hamiltonian dynamics,'' \emph{Handbook
  of markov chain monte carlo}, vol.~2, no.~11, p.~2, 2011.

\bibitem{wainwright2008graphical}
M.~J. Wainwright and M.~I. Jordan, \emph{Graphical models, exponential
  families, and variational inference}.\hskip 1em plus 0.5em minus 0.4em\relax
  Now Publishers Inc, 2008.

\bibitem{adadi2018peeking}
A.~Adadi and M.~Berrada, ``Peeking inside the black-box: a survey on
  explainable artificial intelligence (xai),'' \emph{IEEE access}, vol.~6, pp.
  52\,138--52\,160, 2018.

\bibitem{guidotti2018survey}
R.~Guidotti, A.~Monreale, S.~Ruggieri, F.~Turini, F.~Giannotti, and
  D.~Pedreschi, ``A survey of methods for explaining black box models,''
  \emph{CSUR}, vol.~51, no.~5, pp. 1--42, 2018.

\bibitem{marcinkevivcs2020interpretability}
R.~Marcinkevi{\v{c}}s and J.~E. Vogt, ``Interpretability and explainability: A
  machine learning zoo mini-tour,'' \emph{arXiv preprint arXiv:2012.01805},
  2020.

\bibitem{simonyan2013deep}
K.~Simonyan, A.~Vedaldi, and A.~Zisserman, ``Deep inside convolutional
  networks: Visualising image classification models and saliency maps,''
  \emph{arXiv preprint arXiv:1312.6034}, 2013.

\bibitem{ribeiro2016should}
M.~T. Ribeiro, S.~Singh, and C.~Guestrin, ``" why should i trust you?"
  explaining the predictions of any classifier,'' in \emph{Proceedings of the
  22nd ACM SIGKDD}, 2016, pp. 1135--1144.

\bibitem{sundararajan2017axiomatic}
M.~Sundararajan, A.~Taly, and Q.~Yan, ``Axiomatic attribution for deep
  networks,'' in \emph{ICML}.\hskip 1em plus 0.5em minus 0.4em\relax PMLR,
  2017, pp. 3319--3328.

\bibitem{dhamdhere2018important}
K.~Dhamdhere, M.~Sundararajan, and Q.~Yan, ``How important is a neuron?''
  \emph{arXiv preprint arXiv:1805.12233}, 2018.

\bibitem{jha2020enhanced}
A.~Jha, J.~K. Aicher, M.~R. Gazzara, D.~Singh, and Y.~Barash, ``Enhanced
  integrated gradients: improving interpretability of deep learning models
  using splicing codes as a case study,'' \emph{Genome biology}, vol.~21,
  no.~1, pp. 1--22, 2020.

\bibitem{merrill2019generalized}
J.~Merrill, G.~Ward, S.~Kamkar, J.~Budzik, and D.~Merrill, ``Generalized
  integrated gradients: A practical method for explaining diverse ensembles,''
  \emph{arXiv preprint arXiv:1909.01869}, 2019.

\bibitem{shrikumar2017learning}
A.~Shrikumar, P.~Greenside, and A.~Kundaje, ``Learning important features
  through propagating activation differences,'' in \emph{ICML}.\hskip 1em plus
  0.5em minus 0.4em\relax PMLR, 2017, pp. 3145--3153.

\bibitem{lundberg2017unified}
S.~M. Lundberg and S.-I. Lee, ``A unified approach to interpreting model
  predictions,'' in \emph{Proceedings of the 31st NeurIPS}, 2017, pp.
  4768--4777.

\bibitem{rieger2020simple}
L.~Rieger and L.~K. Hansen, ``A simple defense against adversarial attacks on
  heatmap explanations,'' \emph{arXiv preprint arXiv:2007.06381}, 2020.

\bibitem{tomsett2018failure}
R.~Tomsett, A.~Widdicombe, T.~Xing, S.~Chakraborty, S.~Julier, P.~Gurram,
  R.~Rao, and M.~Srivastava, ``Why the failure? how adversarial examples can
  provide insights for interpretable machine learning,'' in \emph{2018 21st
  FUSION}.\hskip 1em plus 0.5em minus 0.4em\relax IEEE, 2018, pp. 838--845.

\bibitem{dong2017towards}
Y.~Dong, H.~Su, J.~Zhu, and F.~Bao, ``Towards interpretable deep neural
  networks by leveraging adversarial examples,'' \emph{arXiv preprint
  arXiv:1708.05493}, 2017.

\bibitem{madry2017towards}
A.~Madry, A.~Makelov, L.~Schmidt, D.~Tsipras, and A.~Vladu, ``Towards deep
  learning models resistant to adversarial attacks,'' \emph{arXiv preprint
  arXiv:1706.06083}, 2017.

\bibitem{ross2018improving}
A.~S. Ross and F.~Doshi-Velez, ``Improving the adversarial robustness and
  interpretability of deep neural networks by regularizing their input
  gradients,'' in \emph{Thirty-second AAAI}, 2018.

\bibitem{yeh2019fidelity}
C.-K. Yeh, C.-Y. Hsieh, A.~Suggala, D.~I. Inouye, and P.~K. Ravikumar, ``On the
  (in) fidelity and sensitivity of explanations,'' \emph{NeurIPS}, vol.~32, pp.
  10\,967--10\,978, 2019.

\bibitem{bykov2021noisegrad}
K.~Bykov, A.~Hedstr{\"o}m, S.~Nakajima, and M.~M.-C. H{\"o}hne, ``Noisegrad:
  enhancing explanations by introducing stochasticity to model weights,''
  \emph{arXiv preprint arXiv:2106.10185}, 2021.

\bibitem{wenzel2020good}
F.~Wenzel, K.~Roth, B.~S. Veeling, J.~{\'S}wi{a}tkowski, L.~Tran, S.~Mandt,
  J.~Snoek, T.~Salimans, R.~Jenatton, and S.~Nowozin, ``How good is the bayes
  posterior in deep neural networks really?'' \emph{arXiv preprint
  arXiv:2002.02405}, 2020.

\bibitem{mnist}
\BIBentryALTinterwordspacing
Y.~LeCun and C.~Cortes, ``{MNIST} handwritten digit database,'' 2010. [Online].
  Available: \url{http://yann.lecun.com/exdb/mnist/}
\BIBentrySTDinterwordspacing

\bibitem{xiao2017fashion}
H.~Xiao, K.~Rasul, and R.~Vollgraf, ``Fashion-mnist: a novel image dataset for
  benchmarking machine learning algorithms,'' \emph{arXiv preprint
  arXiv:1708.07747}, 2017.

\bibitem{cifar10}
A.~Krizhevsky, V.~Nair, and G.~Hinton, ``The cifar-10 dataset,'' \emph{online:
  http://www. cs. toronto. edu/kriz/cifar. html}, vol.~55, no.~5, 2014.

\bibitem{imagenet_cvpr09}
J.~Deng, W.~Dong, R.~Socher, L.-J. Li, K.~Li, and L.~Fei-Fei, ``{ImageNet: A
  Large-Scale Hierarchical Image Database},'' in \emph{CVPR09}, 2009.

\bibitem{gal2016dropout}
Y.~Gal and Z.~Ghahramani, ``Dropout as a bayesian approximation: Representing
  model uncertainty in deep learning,'' in \emph{ICML}.\hskip 1em plus 0.5em
  minus 0.4em\relax PMLR, 2016, pp. 1050--1059.

\bibitem{smith2018understanding}
L.~Smith and Y.~Gal, ``Understanding measures of uncertainty for adversarial
  example detection,'' \emph{arXiv preprint arXiv:1803.08533}, 2018.

\bibitem{krishnan2020bayesiantorch}
R.~Krishnan and P.~Esposito, ``Bayesian-torch: Bayesian neural network layers
  for uncertainty estimation,'' 2020.

\bibitem{leeintroduction}
J.~Lee, ``Introduction to smooth manifolds (springer, 2012).'' 2012.

\bibitem{montavon2019layer}
G.~Montavon, A.~Binder, S.~Lapuschkin, W.~Samek, and K.-R. M{\"u}ller,
  ``Layer-wise relevance propagation: an overview,'' \emph{Explainable AI:
  interpreting, explaining and visualizing deep learning}, pp. 193--209, 2019.

\bibitem{gilks1995markov}
W.~R. Gilks, S.~Richardson, and D.~Spiegelhalter, \emph{Markov chain Monte
  Carlo in practice}.\hskip 1em plus 0.5em minus 0.4em\relax CRC press, 1995.

\bibitem{betancourt2017conceptual}
M.~Betancourt, ``A conceptual introduction to hamiltonian monte carlo,''
  \emph{arXiv preprint arXiv:1701.02434}, 2017.

\bibitem{metropolis1953equation}
N.~Metropolis, A.~W. Rosenbluth, M.~N. Rosenbluth, A.~H. Teller, and E.~Teller,
  ``Equation of state calculations by fast computing machines,'' \emph{The
  journal of chemical physics}, vol.~21, no.~6, pp. 1087--1092, 1953.

\bibitem{hastings1970monte}
W.~K. Hastings, ``Monte carlo sampling methods using markov chains and their
  applications,'' 1970.

\bibitem{hockney2021computer}
R.~W. Hockney and J.~W. Eastwood, \emph{Computer simulation using
  particles}.\hskip 1em plus 0.5em minus 0.4em\relax crc Press, 2021.

\end{thebibliography}

\newpage
\clearpage
\section{Appendix}

\subsection{Theoretical results}
\label{sec:theoretical_results}

In the following, we summarize the main background material needed for proving Theorem \ref{th:main_thm}.

Theorem 1 in \citep{anders2020fairwashing} is a generalization of the submanifold extension lemma (e.g. Lemma 5.34 in \cite{leeintroduction}). It proves that any function $g$ defined on a submanifold $\mathcal{M}$ can be extended to an embedding manifold $\mathcal{R}$, in such a way that the choice of the derivatives orthogonal to the submanifold is arbitrary. We change the original notation and report it here as Lemma \ref{th:extension_lemma}.

\begin{lemma}
Let $\mathcal{M}\subset \mathcal{R}$ be a $k$-dimensional submanifold embedded in the $d$-dimensional manifold $\mathcal{R}$. Let $V=\sum_{i=k+1}^d v^i \partial_i$ be a conservative vector field along $\mathcal{M}$ which assigns a vector in $T_x\mathcal{M}^\perp$ for each $x\in\mathcal{M}$. For any smooth function $g:\mathcal{M}\to\mathbb{R}$ there exists a smooth extension $F:\mathcal{R}\to\mathbb{R}$ such that 
$$ F|_\mathcal{M} = g,
$$
where $F|_\mathcal{M}$ denotes the restriction of $F$ on the submanifold $\mathcal{M}$, and s.t. the derivative of the extension $F$ is 
$$
\nabla_xF(x)=(\nabla_1 g(x), \ldots, \nabla_k g(x), v^{k+1}(x), \ldots, v^{d}(x))
$$
for all $x\in\mathcal{M}$.
\label{th:extension_lemma}
\end{lemma}

We use Lemma \ref{th:extension_lemma} and Corollary \ref{th:orthogonal_components} to build an extension of the prediction function outside of the data manifold, with suitable derivatives in the orthogonal components.

In the proof of Corollary \ref{th:orthogonal_components} we also leverage results on the global convergence of overparameterized neural networks \citep{du2019gradient, mei2018mean, rotskoff2018neural} to approximate the extension of the prediction function with an infinitely wide neural network. Here we report the Universal Representation Theorem from \cite{rotskoff2018neural} as Lemma \ref{th:global_convergence}, in the limit of infinite training data and infinite number of neurons/ weights (overparameterized network). In doing so, we refer to the notation in Sec. 7 of \cite{carbone2020robustness}.

\begin{lemma}\label{th:global_convergence}
Let $f(x,w)$ be a neural network with differentiable and discriminating units, whose weights are drawn from a measure $\mu(w)$. Let $\tilde{f}(x)$ be a target function observed at points drawn from a data distribution $p(D)$ with support $\mathcal{M}_D \subset \mathbb{R}^d$.
Suppose that:
\begin{itemize}
    \item The input and feature spaces of $f$ are closed Riemannian manifolds.
    \item The data distribution  $p(D)$ is non degenerate.
\end{itemize}
Then, the training loss function is a convex functional of the measure in the space of weights. Moreover, in the infinite data and overparameterized limit, stochastic gradient descent converges to the global minimum $\tilde{f}(x)$, i.e. the loss function is null on the data manifold $\mathcal{M}_D$.
\end{lemma}

Using Lemma \ref{th:extension_lemma} and Lemma \ref{th:global_convergence}, we can prove that it is always possible to select two sets of weights, $w$ and $w'$, such that the orthogonal gradients of the prediction function at a point $x$ on the data manifold are opposite.

\begin{corollary}\label{th:orthogonal_components}
Let $\mathcal{M}_D\subset \mathbb{R}^d$ be an a.e. smooth manifold and let $f(x,w)$ be an infinitely wide Bayesian neural network, at full convergence of the training algorithm on $\mathcal{M}_D$. For any choice of weights $w$ and $x\in\mathcal{M}_D$ there exists a set of weights $w'$ such that $f(\cdot,w)|_{\mathcal{M}_D} = f(\cdot,w')|_{\mathcal{M}_D}$ and
\begin{align}
\nabla_x^\perp f(x,w) = -\nabla_x^\perp f(x,w').
\label{eq:orthogonal_gradients}
\end{align}
\end{corollary}

\begin{proof}
Let $F:\mathbb{R}^d \to \mathbb{R}$ be a smooth function s.t. $F=f(\cdot, w)$.
From \ref{th:extension_lemma} we know that there exists a smooth extension $F'$ of $F|_{\mathcal{M}_D}$ to the embedding space $\mathbb{R}^d$ such that
$\nabla_x^\perp F'(x) = - \nabla_x^\perp f(x)$. Moreover, Lemma \ref{th:global_convergence} holds for $f$ in the limit of infinite weights, thus there exists a set of weights $w'$ such that $f(\cdot,w')=F'$. This proves \ref{eq:orthogonal_gradients}.
\end{proof}

\begin{proof}[Proof of Theorem \ref{th:main_thm}]
As a direct consequence of Corollary \ref{th:orthogonal_components}, under the assumption that the prior distribution $p(w)$ is uninformative, for any choice of weights such that the loss is zero on the data manifold, we obtain the same likelihood function and the same posterior distribution $p(w|D)$. This in particular holds for any pair $w,w'$ constructed according to Corollary \ref{th:orthogonal_components}. Therefore, the orthogonal gradient of the prediction function evaluated on any $x\in\mathcal{M}_D$ vanishes in expectation, i.e.
$$
\mathbb{E}_{p(w\vert D)}\big[\nabla_x^{\perp}f(x,w)\big]=0.
$$
\end{proof}


\subsection{Gradient-based Attacks}
\label{sec:gradient_attacks}

FGSM and PGD attacks are defined as follows:
\begin{itemize}
    \item For a given network $f(\cdot,w)$ FGSM crafts a perturbation in the direction of the greatest loss w.r.t. the input
\begin{align*}
    \tilde{x}=x + \delta \; \text{sgn}  \nabla_x L(x,w),
\end{align*}
where $L$ is the training loss and $\delta$ is the perturbation magnitude ({\it strength} of the attack).
    \item PGD starts from a random perturbation in an $\epsilon\text{-L}_\infty$ ball around the input sample $\tilde{x}_0:=x$. At each step, it performs FSGM with a smaller step size, $\alpha<\epsilon$, and projects the adversary back in the $\epsilon\text{-L}_\infty$ ball
$$\tilde{x}_{t+1}=\text{Proj} \{ \tilde{x}_t+\alpha\cdot \text{sgn}\nabla_x L(\tilde{x}_t,w)\}_{\epsilon, \infty}.
$$ 
The size of the resulting perturbation is smaller than $\epsilon$.
\end{itemize}

\subsection{Saliency Attacks}
\label{sec:saliency_attacks}

In this section we define the adversarial perturbations of the explanations used in our experiments (Sec. \ref{sec:experiments} and Sec. \ref{sec:additional_figures}), whose goal is to alter interpretations without affecting classifications.

\cite{ghorbani2019interpretation} present a variety of attacks against feature importance methods, which iteratively maximize the diversity between original explanations $R(x)$ and perturbed explanations $R(\tilde{x})$. At each step the image is perturbed in the direction of the gradient of a \emph{dissimilarity function} $D(x,\tilde{x})$ 
$$
\tilde{x}_{t+1} = \text{Proj} \{\tilde{x}_t +\alpha \cdot \text{sgn} \nabla_x D(x,\tilde{x}_t) \}_{\epsilon, \infty}.
$$
We leverage two of the proposed techniques, with the following dissimilarity functions:
\begin{itemize}
    \item $D(x, \tilde{x})=\sum_{p \in \mathcal{A}_k} R(\tilde{x}_p)$ for the \emph{target region} attack, where $\mathcal{A}_k$ is a pre-defined region of $k\%$ pixels.
    \item $D(x, \tilde{x})=-\sum_{p\in \text{Top}_k(x)} R(\tilde{x}_p)$ for the \emph{top-k} attack, where $k$ indicates the chosen percentage of most relevant pixels. 
\end{itemize}
In our experiments (Fig. \ref{fig:rules_rob_appendix_ghorbani}) we set $\alpha=0.5$, $k=20$, number of iterations $T=10$ and the region of pixels was chosen randomly.

For each test image $x$ \cite{dombrowski2019explanations} build a targeted perturbation $\tilde{x}$ such that the classification is almost constant but the explanation resembles the target explanation of a randomly chosen test image $\hat{x}\neq x$. Specifically, they optimize the loss function 
$$
||R(\tilde{x})- R(\hat{x})||^2+\gamma \, ||f(\tilde{x})-f(x)||^2
$$
w.r.t. $\tilde{x}$ and clamp $\tilde{x}$ at each iteration to keep the image valid. Additionally, during the optimization phase they replace ReLU activations in the NN with softplus non-linearities
$$ \text{softplus}_\beta(x)=\frac{1}{\beta} \log(1+e^{\beta x})
$$
to avoid the problem of vanishing gradients.
In our tests (Fig. \ref{fig:rules_rob_appendix_dombrowski} (b)) we set $\gamma=1$, iterations $T=10$ and learning rate $lr=0.01$.


\subsection{LRP Rules}
\label{sec:lrp_rules}

\begin{figure}[bt]
	\centering
	\includegraphics[width=0.8\linewidth]{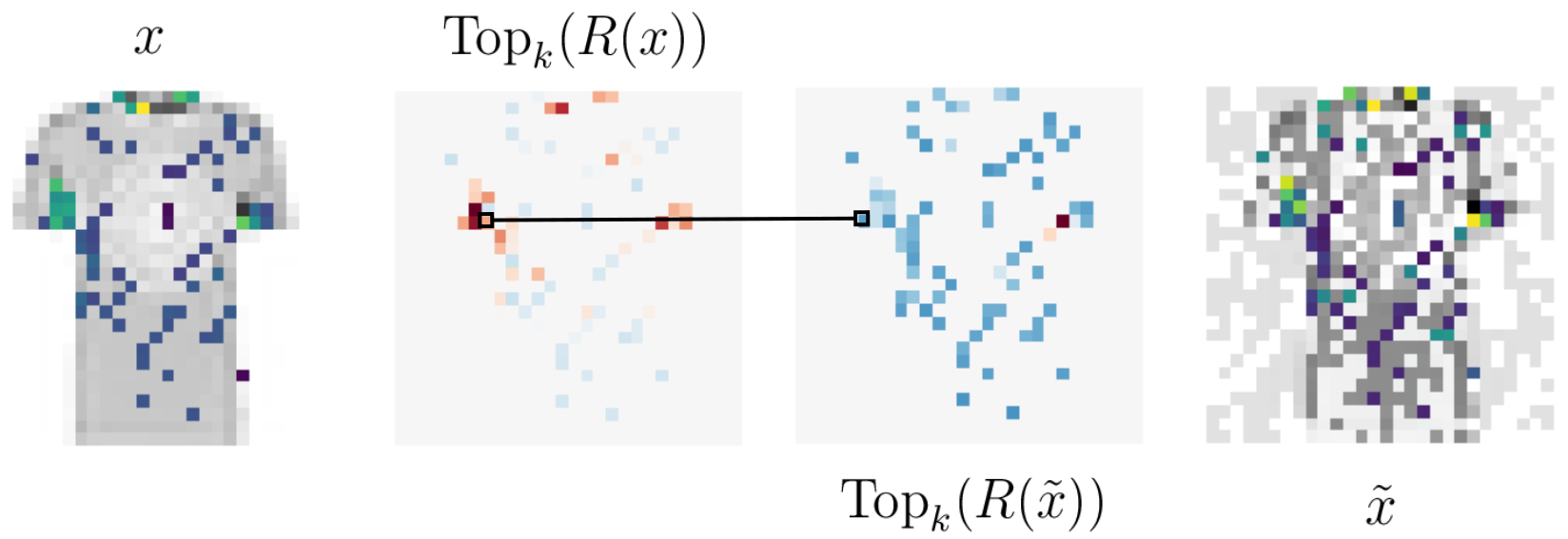}
	\caption{$\text{Top}_{k}$ pixels in an image $x$ from Fashion MNIST dataset and an FGSM adversarial perturbation $\tilde{x}$.}
	\label{fig:topk_rob}
\end{figure}

A practical example of propagation rule is the \emph{Epsilon rule} ($\epsilon$-LRP) \citep{montavon2019layer}. Let $a_j$ be a neuron activation computed at the $j$-th layer during the forward pass. Let $w_{jk}$ be the weight connecting $a_j$ to a subsequent neuron $a_k$, $w_{0k}$ be the bias weight and $a_0=1$. At each step of backpropagation LRP computes the relevance score $R_j$ of the $j$-th activation $a_j$, by backpropagating the relevance scores of all neurons in the subsequent layer. The resulting $\epsilon$-LRP score for a chosen  $\epsilon>0$ amounts to
\begin{align*}
    R_j=\sum_k \frac{a_j w_{jk}}{\epsilon+\sum_{0,j}a_j w_{jk}} R_k.
\end{align*}

The \emph{Gamma rule} ($\gamma$-LRP) favours positive contributions over negative contributions by a factor of $\gamma$. The score for a chosen  $\gamma>0$ is
\begin{align*}
    R_j = \sum_k \frac{a_j\cdot(w_{jk}+\gamma \max(0,w_{jk}))}{\sum_{0,j} a_j \cdot(w_{jk}+\gamma \max(0,w_{jk}))} R_k.
\end{align*}

Finally, the \emph{Alpha-Beta rule} ($\alpha\beta$-LRP) computes
\begin{align*}
    R_j = \sum_k \Bigg(
       &\alpha \cdot \frac{\max(0,a_j w_{jk})}{\sum_{0,j} \max(0,a_j w_{jk})}\\
       - &\beta \cdot \frac{\min(0,a_j w_{jk})}{\sum_{0,j} \min(0,a_j w_{jk})}
    \Bigg) R_k,
\end{align*}
where the conservative property holds for any choice of $\alpha$ and $\beta$ s.t. $\alpha-\beta=1$.

We refer to \cite{montavon2019layer} (Sec 10.2.3) for a complete derivation of the propagation rules listed above within the Deep Taylor Decomposition framework \cite{montavon2017explaining}. Notice that for the $\alpha\beta$-LRP rule this generalization holds only when $\alpha=1$ and $\beta=0$, which are the values used in our experiments.

\subsection{Approximate Inference Methods}
\label{sec:approximate_inference}

We briefly describe the methods used for performing approximate Bayesian inference in our experiments (Sec. \ref{sec:experiments} and \ref{sec:additional_figures}).

\emph{Markov Chain Monte Carlo} (MCMC) \cite{gilks1995markov} is a class of stochastic algorithms that allow to sample from an unknown high-dimensional probability distribution (in our case the posterior $p(w|D)$) by building a Markov chain with the desired distribution as its equilibrium distribution. The efficiency of these methods in high-dimensional spaces is guaranteed by a proper exploration of the \emph{typical set}, which is the region contributing the most to the computation of expectations w.r.t. the target density \cite{betancourt2017conceptual}.  In the limit of infinite samples from the chain, the distribution of the recorded samples exactly matches the posterior distribution.
\emph{Hamiltonian Monte Carlo} (HMC) \cite{neal2011mcmc} is a Markov Chain Monte Carlo technique that combines an approximate Hamiltonian dynamics simulation and a \emph{Metropolis-Hastings} \cite{metropolis1953equation, hastings1970monte} acceptance step. It is designed to generate efficient transitions in the parameter space and to reduce the problems of low acceptance rates and autocorrelation between consecutive samples.
HMC introduces some momentum variables $\rho$ and defines a Hamiltonian
$$
H(\rho, w)= -\log p(\rho|w) -\log p(w)=T(\rho|w)+V(w),
$$
where $p(\rho,w)$ is the joint density $p(\rho|w)p(w)$. Starting from the initial values of $\rho$ and $w$, the system evolves according to Hamilton's equations
\begin{align*}
    \frac{dw}{dt} &= \frac{\partial H}{\partial \rho} = \frac{\partial V}{\partial \rho}\\
    \frac{d \rho}{dt} &= -\frac{\partial H}{\partial w} = -\frac{\partial T}{\partial w}-\frac{\partial V}{\partial w},
\end{align*}
which are solved by means of \emph{Leapfrog integration} \cite{hockney2021computer}.
At each step, Metropolis–Hastings computes the probability of keeping the new samples $(\rho',w')$
$$
\min\{1, \exp(H(\rho, w)-H(\rho',w'))\}.
$$

The second technique we use is \emph{Variational Inference} (VI) \cite{wainwright2008graphical}. It turns Bayesian inference into an optimization problem and provides an analytical approximation $q(w;\nu)\approx p(w|D)$ of the posterior distribution, namely the \emph{variational distribution}, which belongs to a restricted family of known distributions (e.g. Gaussians). The variational parameters $\nu$ are optimized by minimizing the dissimilarity between $p$ and $q$. The measure of similarity is the Kullback-Leiber divergence
$$
\text{KL}(q||p) := -\sum_w q(w;\nu) \log \Big( \frac{p(w|D)}{q(w;\nu)} \Big).
$$
and the minimization problem
$$
\nu^* = {\arg \min}_\nu \text{KL}(q||p)
$$
still requires the computation of the intractable term $p(w|D)$. Notice that the evidence $\log p(D)$ is constant with respect to $\nu$ and satisfies the equality
$$
\log p(D) = \text{KL}(q||p) + \mathbb{E}_q [\log p(w,D)-\log q(w;\nu)].
$$
Therefore, the optimization problem above is equivalent to the minimization of the \emph{Evidence Lower Bound} (ELBO) loss
$$
\text{ELBO}(\nu)=\mathbb{E}_q [\log p(w,D)-\log q(w;\nu)].
$$
The first term in the ELBO loss encourages $q$ to place its probability mass on the MAP estimate, while the second favours entropy on the mass, to avoid its concentration in a single location.

\subsection{Computational Resources}
\label{sec:comput_resources}
Simulations were run on a machine with 36 cores, Intel(R) Xeon(R) Gold 6140 CPU @ 2.30GHz processors and 192GB of RAM.

\newpage
\subsection{Architectures and Hyperparameters}
\label{sec:architectures}

\begin{table}[h]
\caption{Learnable layers and corresponding indexes in pytorch for the convolutional architecture. Hidden size is reported in Tab. \ref{table:vi_hyperparams} and Tab. \ref{table:hmc_hyperparams}.}
\small
\centering
\label{tab:conv_architecture}
\begin{tabular}{c|c|c}
    \toprule
    \bf Idx & \bf Layer & \bf Parameters\\
    \midrule
    $0$ & 2D Conv. &
    \def\arraystretch{1}\begin{tabular}{@{}c@{}} \texttt{in\_channels} = $784$ \\
    \texttt{out\_channels} = $32$ \\ \texttt{kernel\_size} = $5$ \\
    \end{tabular}\\
    & & \\
    $3$ & 2D Conv. & 
    \def\arraystretch{1}\begin{tabular}{@{}c@{}} \texttt{in\_channels} = $32$ \\
    \texttt{out\_channels} = hidden size \\ \texttt{kernel\_size} = $5$ \\
    \end{tabular}\\
    & & \\
    $7$ & F.c. &
    \def\arraystretch{1}\begin{tabular}{@{}c@{}} \texttt{in\_features} = hidden size \\
    \texttt{out\_features} = $10$\\
    \end{tabular}\\
    \bottomrule
\end{tabular}
\end{table}

\begin{table}[h]
\caption{Learnable layers and corresponding indexes in pytorch for the fully connected architecture. Hidden size is reported in Tab. \ref{table:vi_hyperparams} and Tab. \ref{table:hmc_hyperparams}.}
\small
\centering
\begin{tabular}{c|c|c}
    \toprule
    \bf Idx & \bf Layer & \bf Parameters\\
    \midrule
    $1$ & F. c. &
    \def\arraystretch{1}\begin{tabular}{@{}c@{}} \texttt{in\_features} = $784$ \\
    \texttt{out\_features} = hidden size\\
    \end{tabular}\\
    & & \\
    $3$ & F. c. &     \def\arraystretch{1}\begin{tabular}{@{}c@{}} \texttt{in\_features} = hidden size \\
    \texttt{out\_features} = hidden size\\
    \end{tabular}\\ 
    & & \\
    $5$ & F. c. &
    \def\arraystretch{1}\begin{tabular}{@{}c@{}} \texttt{in\_features} = hidden size \\
    \texttt{out\_features} = $10$\\
    \end{tabular}\\ 
    \bottomrule
\end{tabular}
\label{tab:fc_architecture}
\end{table}

\begin{table}[h]
\caption{Hyperparameters for BNNs trained with VI.}
\small
\centering
\begin{tabular}{p{3cm}||c|c}
 \toprule
 \bf Dataset & \bf MNIST & \bf Fashion MNIST \\
 \midrule
 Training inputs & 60k & 60k \\
 Hidden size & 512 & 1024 \\
 Nonlinear activations & Leaky ReLU & Leaky ReLU\\
 Architecture & Convolutional & Convolutional \\
 Training epochs & 5 & 15\\
 Learning rate & 0.01 & 0.001 \\
 \bottomrule
\end{tabular}
\label{table:vi_hyperparams}
\end{table}

\begin{table}[htb]
\caption{Hyperparameters for BNNs trained with HMC.}
\small
\centering
\begin{tabular}{ p{3cm}||c|c}
 \toprule
 \bf Dataset & \bf MNIST & \bf Fashion MNIST \\
 \midrule
 Training inputs & 60k & 60k \\
 Hidden size & 512 & 1024 \\
 Nonlinear activation &  Leaky ReLU &  Leaky ReLU\\
 Architectures & Fully connected & Fully connected \\
 Warmup samples & 100 & 100\\
 Numerical integrator stepsize & 0.5 & 0.5\\
 Numerical integrator n. of steps & 10 & 10\\
 \bottomrule
\end{tabular}
\label{table:hmc_hyperparams}
\end{table}


\subsection{Additional Experiments}
\label{sec:additional_figures}

This section presents some more experiments performed on MNIST, Fashion MNIST and CIFAR-10 datasets, using BNNs trained with HMC and VI. Layer indexes refer to the learnable layers in each architecture. Parameters for computing the LRP heatmaps are set to: $\epsilon=0.1, \gamma=0.1, \alpha=1, \beta=0$.

\begin{figure}[htb]
	\centering
	\begin{subfigure}{0.5\linewidth}
        \centering	\includegraphics[width=\linewidth]{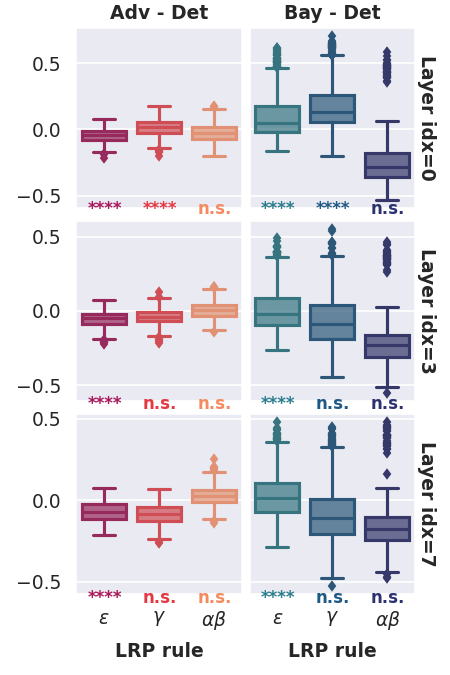}
        \caption{}
    \end{subfigure}%
    \hfill
	\begin{subfigure}{0.5\linewidth}
    	\centering	   \includegraphics[width=\linewidth]{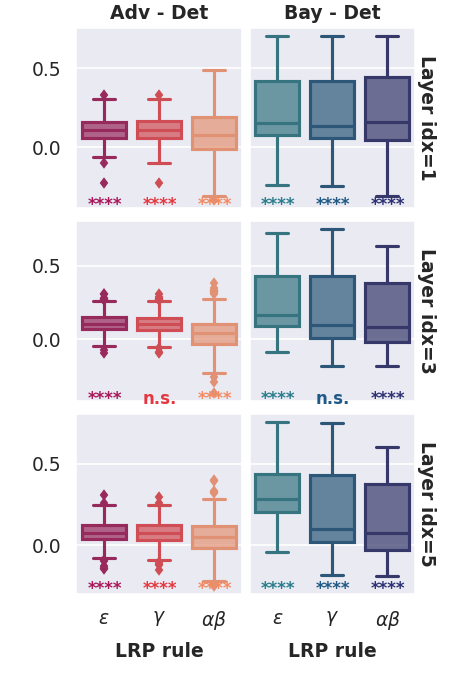}
		\caption{}
    \end{subfigure}
	\caption{LRP robustness differences for FGSM attacks computed on $500$ test points from MNIST dataset using Epsilon, Gamma and Alpha-Beta rules on the $\text{Top}_{20}$ pixels. NNs in (a) have a convolutional architecture, while NNs in (b) have a fully connected architecture. The BNN in (a) is trained with VI and the BNN in (b) is trained with HMC; Both are tested using $100$ posterior samples. 
	}
    \label{fig:rules_rob_appendix_1}
\end{figure}

\begin{figure}[htb]
	\centering
	\begin{subfigure}{0.5\linewidth}
        \centering	\includegraphics[width=\linewidth]{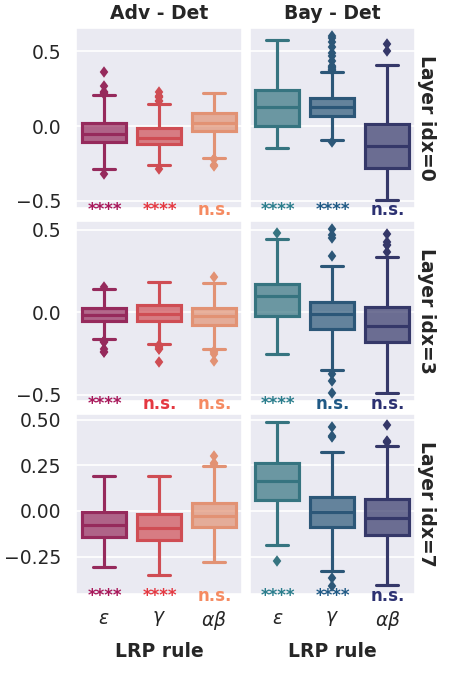}
        \caption{}
    \end{subfigure}%
    \hfill
	\begin{subfigure}{0.5\linewidth}
    	\centering	   \includegraphics[width=\linewidth]{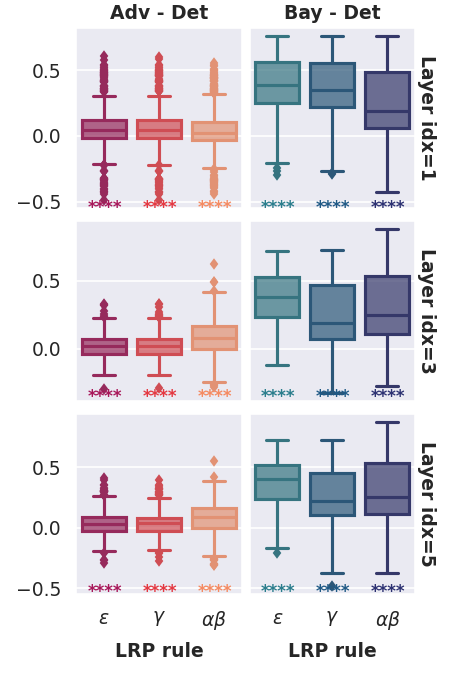}
		\caption{}
    \end{subfigure}
	\caption{LRP robustness differences for FGSM attacks computed on $500$ test points from Fashion MNIST dataset using Epsilon, Gamma and Alpha-Beta rules on the $\text{Top}_{20}$ pixels. NNs in (a) have a convolutional architecture, while NNs in (b) have a fully connected architecture. The BNN in (a) is trained with VI and the BNN in (b) is trained with HMC; Both are tested using $100$ posterior samples. 
	}
    \label{fig:rules_rob_appendix_2}
\end{figure}

\begin{figure}[htb]
	\centering
	\begin{subfigure}{0.5\linewidth}
        \centering	\includegraphics[width=\linewidth]{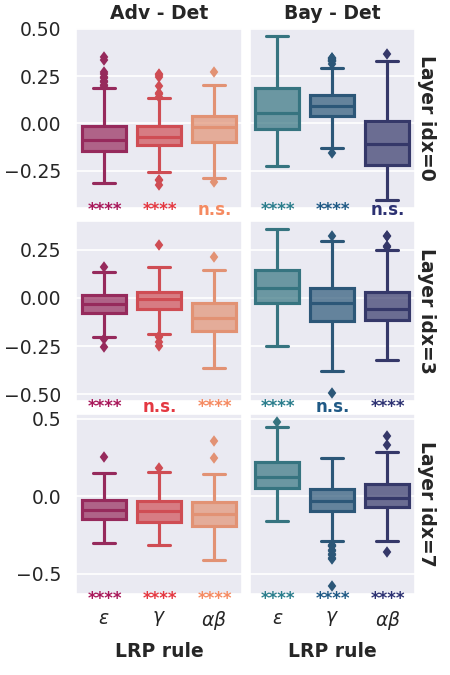}
        \caption{}
    \end{subfigure}%
    \hfill
	\begin{subfigure}{0.5\linewidth}
    	\centering	   \includegraphics[width=\linewidth]{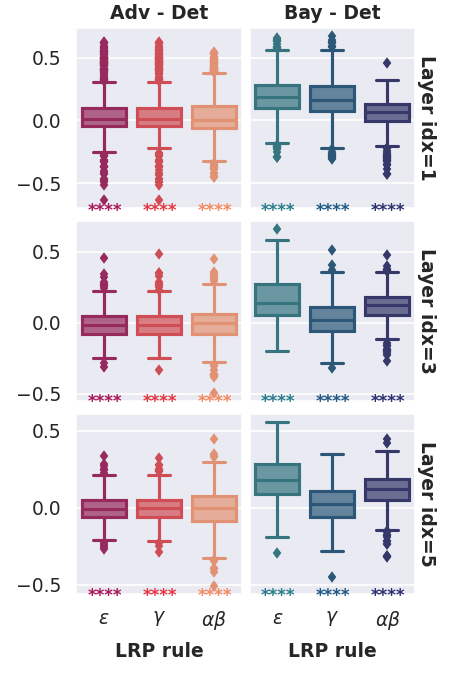}
		\caption{}
    \end{subfigure}
	\caption{LRP robustness differences for PGD attacks computed on $500$ test points from Fashion MNIST dataset using Epsilon, Gamma and Alpha-Beta rules on the $\text{Top}_{20}$ pixels. NNs in (a) have a convolutional architecture, while NNs in (b) have a fully connected architecture. The BNN in (a) is trained with VI and the BNN in (b) is trained with HMC; Both are tested using $100$ posterior samples. 
	}
    \label{fig:rules_rob_appendix_3}
\end{figure}

\begin{figure}[htb]
	\centering	 \includegraphics[width=0.5\linewidth]{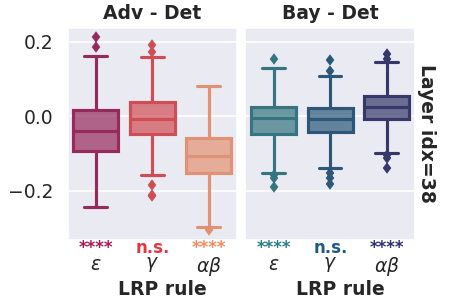}
	\caption{LRP robustness differences for PGD attacks computed on $500$ test points from CIFAR-10 dataset using Epsilon, Gamma and Alpha-Beta rules on the $\text{Top}_{20}$ pixels. NNs have a ResNet20 architecture from \texttt{bayesian\_torch} library \cite{krishnan2020bayesiantorch}. The BNN is trained with VI and tested using $100$ posterior samples. 
	}
	\label{fig:rules_rob_appendix_cifar}
\end{figure}

\begin{figure}[htb]
	\centering
	\begin{subfigure}{0.5\linewidth}
        \centering	\includegraphics[width=\linewidth]{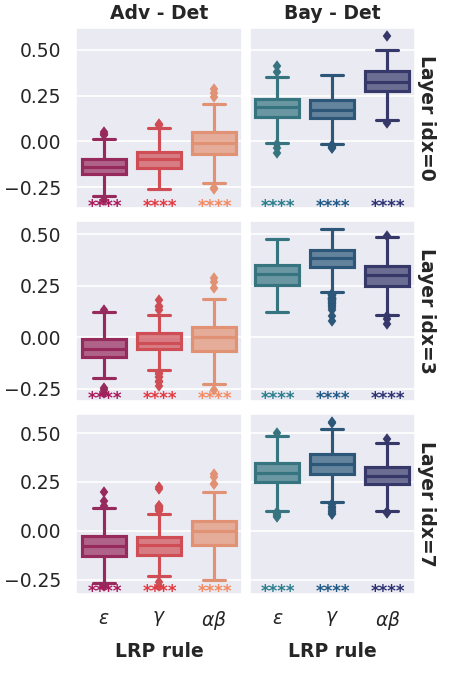}
        \caption{}
    \end{subfigure}%
    \hfill
	\begin{subfigure}{0.5\linewidth}
    	\centering	   \includegraphics[width=\linewidth]{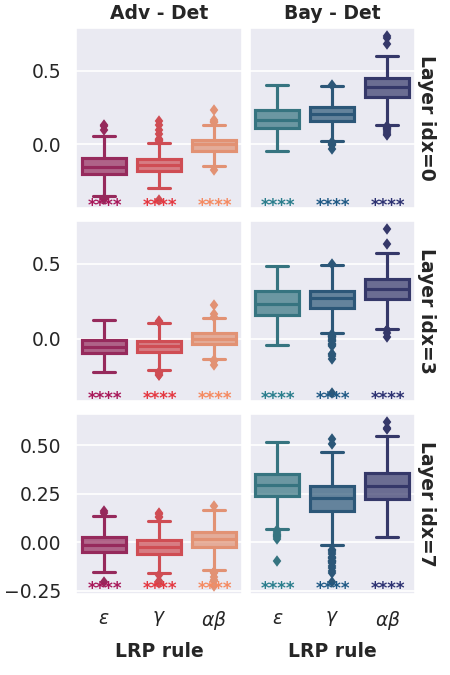}
		\caption{}
    \end{subfigure}
	\caption{LRP robustness differences for top-k (a) and target region (b) attacks \cite{ghorbani2019interpretation} (Sec. \ref{sec:saliency_attacks}) computed on $500$ test points from MNIST (a) and Fashion MNIST (b) datasets using Epsilon, Gamma and Alpha-Beta rules on the $\text{Top}_{20}$ pixels. NNs in (a) have a convolutional architecture, while NNs in (b) have a fully connected architecture. BNNs are trained with VI and tested using $100$ posterior samples. 
	}
    \label{fig:rules_rob_appendix_ghorbani}
\end{figure}

\begin{figure}[htb]
	\centering
	\includegraphics[width=0.5\linewidth]{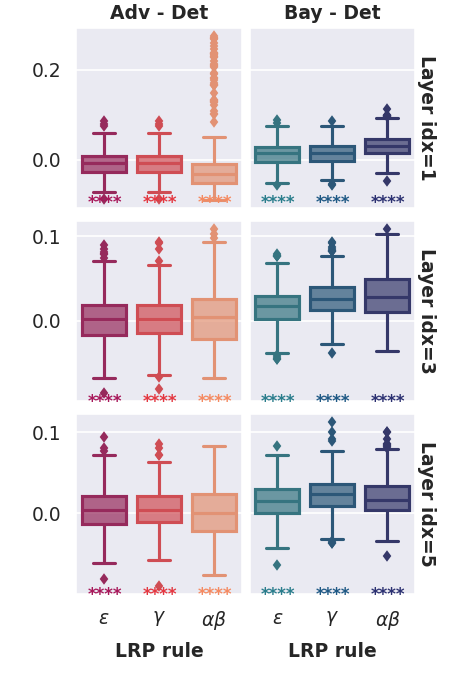}
	\caption{LRP robustness differences for target beta attacks \cite{dombrowski2019explanations} (Sec. \ref{sec:saliency_attacks}) computed on $500$ test points from MNIST dataset using Epsilon, Gamma and Alpha-Beta rules on the $\text{Top}_{60}$ pixels. NNs have a fully connected architecture (Tab. \ref{tab:fc_architecture} in the Appendix). The BNN is trained with HMC and tested using $100$ posterior samples. 
	}
	\label{fig:rules_rob_appendix_dombrowski}
\end{figure}

In Fig. \ref{fig:rules_rob_appendix_1}-\ref{fig:rules_rob_appendix_dombrowski} we also compare the robustness distributions for adversarially trained and Bayesian Neural Networks using Mann-Whitney U rank test. The asterisk notation (Tab. \ref{table:stat_test_symbols}) indicates statistically significant p-values in favour of the alternative hypothesis that the the distribution from the adversarially trained model is stochastically lower than the distribution from the Bayesian model. For significant p-values the Bayesian model is significantly more robust than the deterministic model. 

\begin{table}[htb]
\caption{Asterisk notation for Mann-Whitney test. 
}
\small
\centering
\begin{tabular}{l|c}
 \toprule
 \bf p-value & \bf symbol \\
 \midrule
    $p> 0.05$ & n.s. \\
    $p\leq 0.05$ & * \\
    $p\leq 0.01$ & ** \\
    $p\leq 0.001$ & *** \\
    $p\leq 0.0001$ & ***** \\
 \bottomrule
\end{tabular}
\label{table:stat_test_symbols}
\end{table}

\begin{figure}[htb]
	\centering
	\begin{subfigure}{0.5\linewidth}
        \centering	\includegraphics[width=\linewidth]{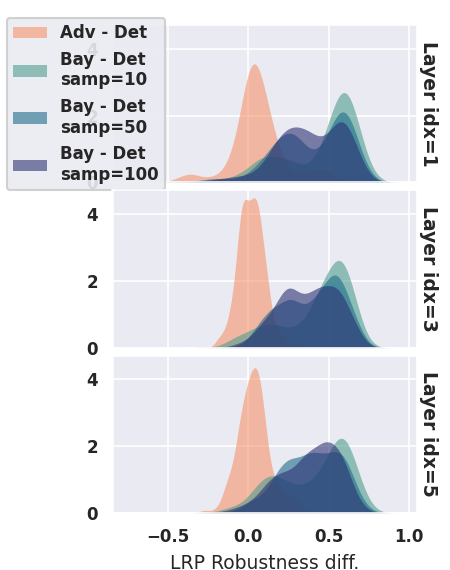}
        \caption{}
    \end{subfigure}%
    \hfill
	\begin{subfigure}{0.5\linewidth}
    	\centering	\includegraphics[width=\linewidth]{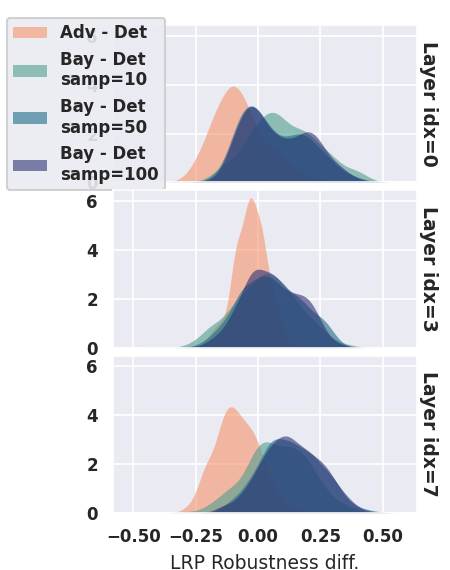}
    	\caption{}
    \end{subfigure}
    \caption{LRP robustness differences for FGSM (a) and PGD (b) attacks computed on $500$ test points from Fashion MNIST dataset using the Epsilon rule. NNs in (a) have a fully connected architecture
    and the BNN is trained with HMC. NNs in (b) have a convolutional architecture  and the BNN is trained with VI. BNNs are tested using an increasing number of samples ($10,50,100$). 
    }
    \label{fig:rob_diff_appendix_1}
\end{figure}

\begin{figure}[htb]
	\centering
	\begin{subfigure}{0.5\linewidth}
        \centering	\includegraphics[width=\linewidth]{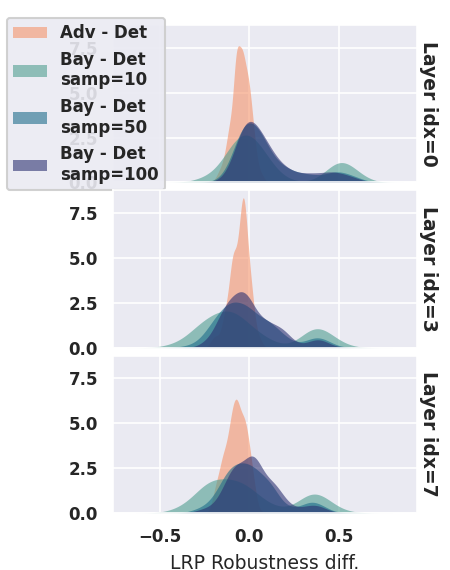}
        \caption{}
    \end{subfigure}%
    \hfill
	\begin{subfigure}{0.5\linewidth}
    	\centering	\includegraphics[width=\linewidth]{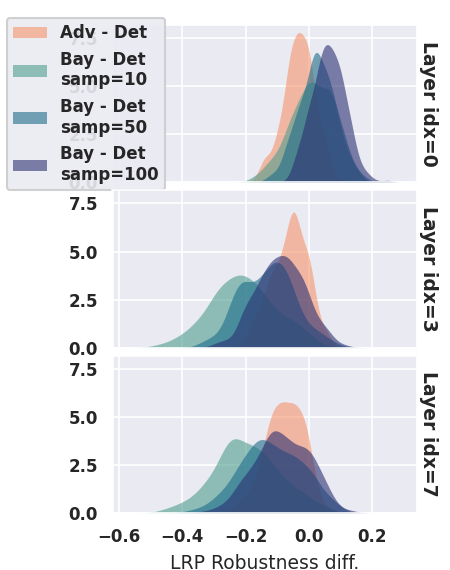}
    	\caption{}
    \end{subfigure}
    \caption{LRP robustness differences for FGSM (a) and PGD (b) attacks computed on $500$ test points from MNIST dataset using the Epsilon rule. NNs have a convolutional architecture. BNNs are trained with VI and tested using an increasing number of samples ($10,50,100$). Layer indexes refer to the learnable layers in the architectures.}
    \label{fig:rob_diff_appendix_2}
\end{figure}

\begin{figure}[htb]
	\centering
	\begin{subfigure}{0.5\linewidth}
        \centering	\includegraphics[width=\linewidth]{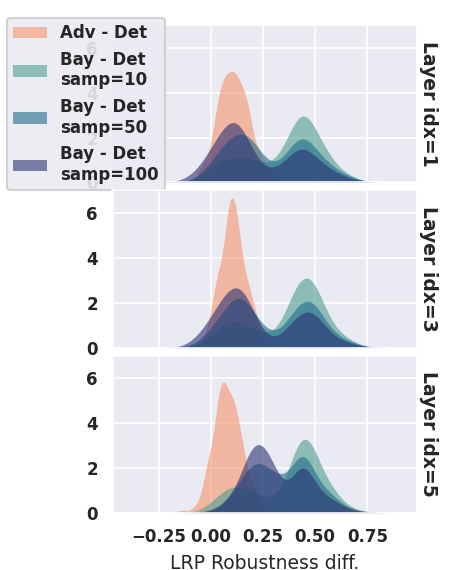}
        \caption{}
    \end{subfigure}%
    \hfill
	\begin{subfigure}{0.5\linewidth}
    	\centering	\includegraphics[width=\linewidth]{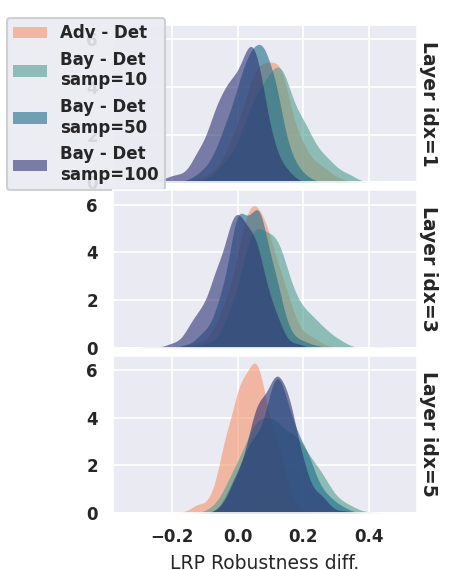}
    	\caption{}
    \end{subfigure}
    \caption{LRP robustness differences for FGSM (a) and PGD (b) attacks computed on $500$ test points from MNIST dataset using the Epsilon rule. NNs have a fully connected architecture. BNNs are trained with HMC and tested using an increasing number of samples ($10,50,100$). 
    }
    \label{fig:rob_diff_appendix_3}
\end{figure}

\begin{figure}[htb]
    \centering
    	\begin{subfigure}{0.5\linewidth}
            \centering	\includegraphics[width=\linewidth]{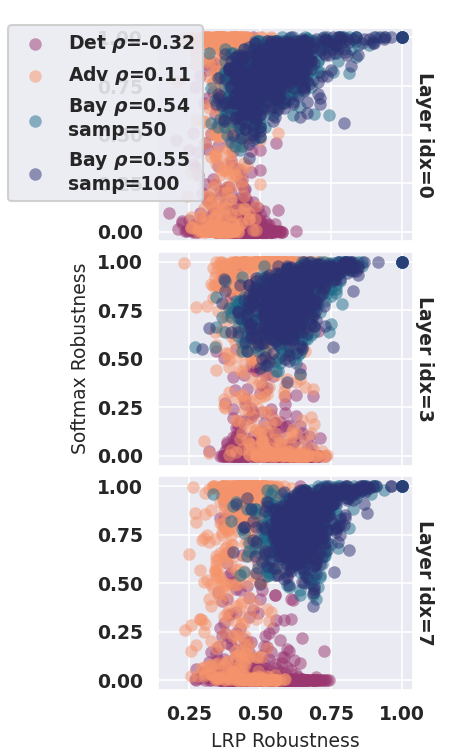}
            \caption{}
        \end{subfigure}%
        \hfill
    	\begin{subfigure}{0.5\linewidth}
        	\centering	\includegraphics[width=\linewidth]{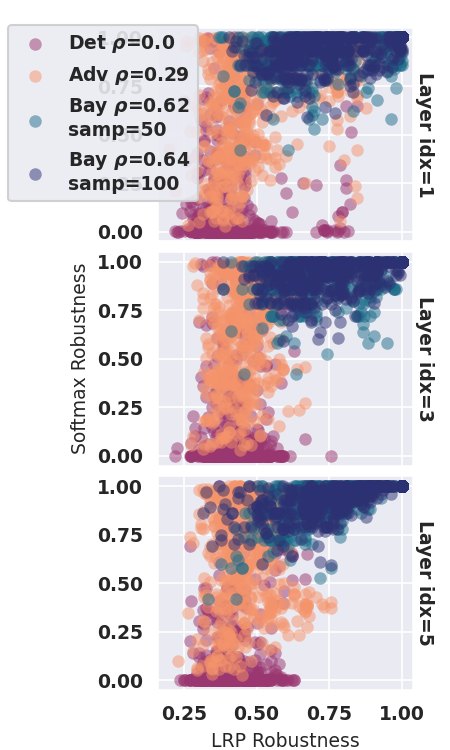}
        	\caption{}
        \end{subfigure}
	\caption{LRP vs softmax robustness of deterministic, adversarially trained and Bayesian NNs trained on Fashion MNIST dataset and tested against FGSM attacks. $\rho$ denotes the correlation coefficient. LRP Robustness is computed with the Epsilon rule on the $20\%$ most relevant pixels. BNNs are trained with VI (a) and HMC (b) and are tested using an increasing number of samples ($50,100$). 
	}
    \label{fig:scatterplot_appendix}
\end{figure}

\end{document}